\documentclass[10pt,twocolumn,letterpaper]{article}
\usepackage{pdfpages}
\usepackage{multido}

\usepackage{cvpr}              

\usepackage[accsupp]{axessibility} 

\usepackage{graphicx}
\usepackage{amsmath}
\usepackage{amssymb}
\usepackage{booktabs}
\usepackage{multirow}
\usepackage{xcolor}

\usepackage[pagebackref,breaklinks,colorlinks]{hyperref}

\newcommand\blfootnote[1]{%
  \begingroup
  \renewcommand\thefootnote{}\footnote{#1}%
  \addtocounter{footnote}{-1}%
  \endgroup
}

\usepackage[capitalize]{cleveref}
\crefname{section}{Sec.}{Secs.}
\Crefname{section}{Section}{Sections}
\Crefname{table}{Table}{Tables}
\crefname{table}{Tab.}{Tabs.}

\definecolor{somegray}{rgb}{0.5, 0.5, 0.5}
\newcommand{\darkgrayed}[1]{\textcolor{somegray}{#1}}
\makeatletter
\newcommand*\titleheader[1]{\gdef\@titleheader{#1}}
\AtBeginDocument{%
  \let\st@red@title\@title
  \def\@title{%
    \vskip-3em
    \bgroup\normalfont\large\centering\@titleheader\par\egroup
    \vskip1.5em\st@red@title}
}
\makeatother

\titleheader{\darkgrayed{This paper has been accepted for publication at the \\
IEEE Conference on Computer Vision and Pattern Recognition (CVPR), New Orleans, 2022.
\copyright IEEE}}

\title{Open Challenges in Deep Stereo: the Booster Dataset}

\begin{document}

\author{ Pierluigi Zama Ramirez$^*$  \hspace*{1cm} Fabio Tosi$^*$ \hspace*{1cm}  
Matteo Poggi$^*$\\ Samuele Salti \hspace*{1cm} Stefano Mattoccia  \hspace*{1cm} Luigi Di Stefano \\
CVLAB, Department of Computer Science and Engineering (DISI)\\
University of Bologna, Italy\\
{\tt\small \{pierluigi.zama, fabio.tosi5, m.poggi\}@unibo.it}
}

\twocolumn[{
\renewcommand\twocolumn[1][]{#1}
\maketitle
\begin{center}
    \vspace{-1cm}
    \begin{tabular}{c c c c}
        \includegraphics[width=0.18\textwidth]{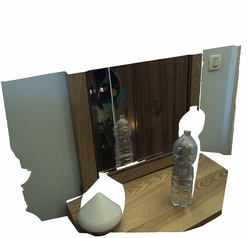} &
        \includegraphics[width=0.18\textwidth]{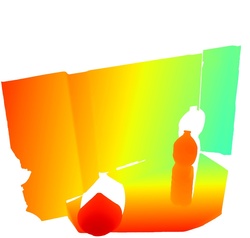} &
        \includegraphics[width=0.18\textwidth]{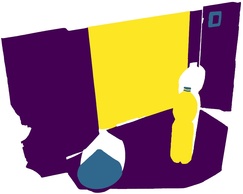} &
        \includegraphics[width=0.18\textwidth]{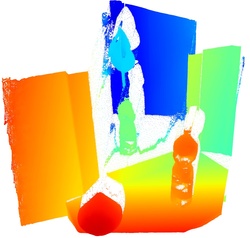} \\
        \small \textbf{(a)} & \small \textbf{(b)} & \small \textbf{(c)} & \small \textbf{(d)} \\
    \end{tabular}
    \label{fig:teaser}
\end{center}
\vspace{-0.2cm}
\small \hypertarget{fig:teaser}{Fig. 1.} \textbf{A scene in the Booster dataset.} We collect images in a variety of indoor environments featuring challenging objects such as the mirror shown in (a). We provide dense ground-truth disparities (b) and
segmentation masks that identify the most challenging materials (c). As state-of-the-art deep stereo networks \cite{lipson2021raft} struggle on these scenes (d), our benchmark highlights the open challenges in deep stereo. 
\vspace{0.5cm}
}]

\maketitle

\begin{abstract}
    We present a novel high-resolution and challenging stereo dataset framing indoor scenes annotated with dense and accurate ground-truth disparities. 
    Peculiar to our dataset is the presence of several specular and transparent surfaces, i.e. the main causes of failures for state-of-the-art stereo networks.
    Our acquisition pipeline leverages a novel deep space-time stereo framework which allows for easy and accurate labeling with sub-pixel precision.
    We release a total of 419 samples collected in 64 different scenes and annotated with dense ground-truth disparities. Each sample include a high-resolution pair (12 Mpx) as well as an unbalanced pair (Left: 12 Mpx, Right: 1.1 Mpx). Additionally, we provide manually annotated material segmentation masks and 15K unlabeled samples.  We evaluate state-of-the-art deep networks based on our dataset, highlighting their limitations in addressing the open challenges in stereo and drawing hints for future research. 

\end{abstract}

\section{Introduction}
\label{sec:intro}

\blfootnote{$^*$ Joint first authorship.}Depth estimation from images has long been deemed a favourable alternative compared to expensive and intrusive active sensors. 
Among several image-based approaches, stereo vision \cite{scharstein2002taxonomy,poggi2021synergies} is arguably the most popular and heavily researched technique. In the years, huge progresses have been made in this field, also thanks to the availability of challenging stereo benchmarks \cite{Geiger2012CVPR,Menze2015CVPR,scharstein2014high,schoeps2017cvpr} where the community competes for the higher ranks. 
Moreover, the abundance of stereo images paved the way for deep learning to succeed also in this field \cite{zbontar2016stereo,mayer2016large,Kendall_2017_ICCV}. Indeed, by browsing the most popular benchmarks, one can notice how nowadays all the top-ranking proposals consist in end-to-end deep networks that can reach sub-pixel precision in most cases. Just to name a few, KITTI 2012 and 2015 \cite{Geiger2012CVPR,Menze2015CVPR} or ETH3D \cite{schoeps2017cvpr} seem solved, with top entries achieving average error rates near to 1\%. Should this evidence suggest that, thanks to deep learning, stereo vision is a solved problem? As shown in Fig. \hyperref[fig:teaser]{1}, we believe that this is definitely not the case and, rather, it is time for the community to focus on the \textbf{open-challenges} left unsolved in the field. In particular, we identify two of such challenges, namely i) non-Lambertian surfaces and ii)  high-resolution images.

As for non-Lambertian reflectivity,  a variety of materials and surfaces still represent a hard challenge to most computer vision methodologies and to deep stereo alike. Specifically, matching pixels dealing with transparent or specular surfaces is extremely difficult and may consist in an inherently  ill-posed problem in many cases. Yet, we reckon that objects with such properties are almost absent or unlabeled in most stereo benchmarks, except for KITTI 2015, where cars have been replaced with CAD models providing supervision on some specular/transparent surfaces on cars. As reported in the KITTI 2015 online benchmark, deep learning has the potential to tackle this challenge as well, if properly annotated samples are available.

Concerning the second challenge, when considering  higher-resolution images, for instance in the Middlebury 2014 benchmark \cite{scharstein2014high}, we can notice in general higher errors. These are caused by the much larger image dimensions (and thus disparity range) and, consequently, by a larger number of occluded and untextured pixels in the images framed in this dataset. Besides, processing images at high resolution sets forth  computational complexity issues, in particular when deploying deep networks. Indeed, most of the entries in the Middlebury benchmark can only process input images downsampled to half or quarter of the original 6 Mpx resolution.
Moreover, an additional challenge emerges due the peculiar  camera setup featured by modern smartphones, typically equipped with both a high resolution and a much lower resolution image sensors. In such a setting, one may wish to recover a high resolution depth map despite the different resolution of the input pair, i.e., solve an \textit{unbalanced} stereo problem.  However, such a research direction has been only barely explored so far \cite{liu2020visually,aleotti2021neural}.

To this aim, in this paper we present a novel high-resolution challenging stereo benchmark. Each image in our dataset, collected in indoor environments, features a set of objects and surfaces that are either specular or transparent, as well as very large untextured regions. To accurately annotate each collected sample, we implement a novel \textit{deep} space-time stereo pipeline \cite{davis2003spacetime} which combines disparity estimates computed from multiple static images -- up to 100 -- acquired under a variety of texture patterns projected onto the scene from different directions and after having carefully painted all non-Lambertian surfaces. Peculiar to our pipeline is the use of a state-of-the-art, pretrained deep network \cite{lipson2021raft} to compute the individual disparity maps accumulation through time within the space-time framework. Furthermore, a final careful manual cleaning is carried out to remove outliers/artefacts and ensure high-quality disparity labels.
We point out that for some non-Lambertian surfaces it might be possible to provide multiple depth ground-truths: for instance, for transparent surfaces we might provide both depths for the surface itself and the objects seen through the surface. Yet, in our dataset we provide depth labels for the closest surfaces only, thereby enabling evaluation and training of stereo methods designed to return a single depth prediction per pixel. As such, our dataset mainly addresses scenarios dealing with autonomous driving, obstacle avoidance and robotic manipulation, while being less amenable to applications such as AR and novel view synthesis.
The main contributions of our paper are:

$\bullet$ We propose a novel dataset consisting of both high-resolution as well as unbalanced stereo pairs featuring a large collection of labeled non-Lambertian objects. In particular, we have acquired a total of 64 scenes under different illuminations, yielding 419 \textit{balanced} stereo pairs at 12 Mpx and 419 \textit{unbalanced} pairs, each consisting in a 12 Mpx and 1.1 Mpx image. The latter setup provides the \textbf{first-ever} dataset for unbalanced stereo matching, as prior work  is limited to simulation experiments \cite{liu2020visually,aleotti2021neural}.
In both setups, samples are annotated with dense ground-truth disparities and grouped into 228 training images and 191 test images -- for which ground-truth is withheld.

$\bullet$ Data annotation is performed in a semi-automatic manner based on a novel deep space-time stereo framework, which enables to deploy modern stereo networks \cite{lipson2021raft} within the well-known space-time stereo framework  \cite{davis2003spacetime}. 

$\bullet$ Alongside with ground-truth disparities, we provide manually annotated segmentation maps that identify and rank the hard-to-match materials based on specularity and transparency. This is conducive to focus on the open-challenges addressed in this paper when analyzing the behaviour of state-of-the-art networks. Moreover, we provide an additional set of 15K raw pairs, both in balanced and unbalanced settings, to encourage the development of weakly-supervised solutions to the open challenges in stereo.

$\bullet$ We evaluate the prominent state-of-the-art stereo networks \cite{chang2018psmnet,zhang2019ga,yang2019hierarchical,cheng2020hierarchical}, as trained by their authors, on the test split of our dataset. The experimental findings highlight the open-challenges that need to be faced by the stereo community and provide hints on possible future research directions. 
Our \underline{B}enchmark \underline{o}n \underline{o}pen-challenges in \underline{ster}eo (\textbf{Booster}) is available at \url{https://cvlab-unibo.github.io/booster-web/}.

\section{Related Work}

We briefly review the literature relevant to our work.

\textbf{Traditional and Deep Stereo.} For years, most algorithms have been developed following a common pipeline sketched in \cite{scharstein2002taxonomy}, starting with matching cost computation and successive optimization strategies. Among the vast literature on traditional algorithms \cite{di2004fast,yoon2006adaptive,yang2012non,de2011linear,hosni2012fast}, Semi-Global Matching (SGM) \cite{hirschmuller2007stereo} is by far the most popular.
With the advent of deep learning, the first research efforts focused on formulating the individual steps of the conventional pipeline \cite{scharstein2002taxonomy} as learnable neural networks, \eg matching cost computation \cite{zbontar2016stereo,Chen_2015_ICCV,luo2016efficient}, optimization \cite{seki2016patch,seki2017sgm-net} and refinement \cite{batsos2018recresnet,gidaris2017detect,aleotti2021neural}.
Then, end-to-end deep stereo networks rapidly gained the main stage \cite{mayer2016large,Kendall_2017_ICCV,Pang_2017_ICCV_Workshops}, thanks to the top-positions achieved on the KITTI 2012 \cite{Geiger2012CVPR} and 2015 \cite{Menze2015CVPR} benchmarks.

\begin{figure*}[t]
    \centering
    \includegraphics[width=0.8\textwidth]{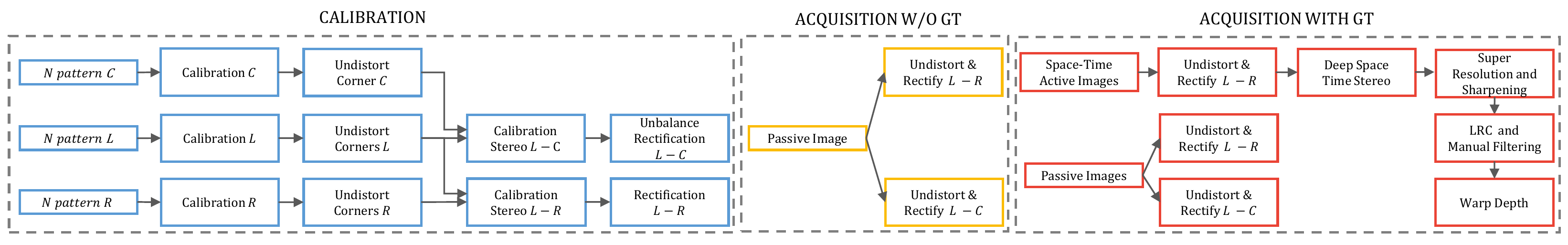}
    \vspace{-0.25cm}\caption{\textbf{Dataset acquisition overview.} Our dataset acquisition procedure can be divided into 3 main parts. Left (blue): initial calibration of our trinocular rig and the two stereo systems $L-C$ and $L-R$. Middle (yellow): image acquisition without ground-truth. Right (red): acquisition with ground-truth.}
    \label{fig:pipeline}
\end{figure*}

This research direction produced a large variety of deep stereo architectures \cite{Liang_2018_CVPR,chang2018psmnet,zhang2019ga,yang2019hierarchical,cheng2020hierarchical,song2018edgestereo,yang2018segstereo,dovesi2020real,Tonioni_2019_CVPR},
as surveyed in \cite{poggi2021synergies}, as well as investigations on self-supervised learning strategies \cite{Tonioni_2017_ICCV,tonioni2020unsupervised,Zhou_2017_ICCV,wang2019unos,lai19cvpr,aleotti2020reversing,Poggi2021continual}, zero-shot generalization across datasets \cite{cai2020matchingspace,zhang2019domaininvariant,aleotti2021neural} and, more recently, unbalanced stereo setups \cite{liu2020visually,aleotti2021neural}.

\textbf{Stereo benchmarks.} Among the factors behind the intensive research on stereo vision, the increasing availability of datasets and benchmarks plays a crucial role. For the first decades, the dataset were limited to few dozen samples, acquired in controlled environments and mostly made available by the Middlebury benchmark \cite{scharstein2002taxonomy,scharstein2003high,hirschmuller2007evaluation,scharstein2007learning}. In the `10s, more and more stereo datasets appeared, starting with KITTI 2012 \cite{geiger2010efficient} and 2015 \cite{Menze2015CVPR}, collected in driving environments and annotated by means of a Velodyne LiDAR sensor, then Middlebury 2014 \cite{scharstein2014high}, framing indoor environments at up to 6 Mpx and annotated through pattern projection, and ETH3D \cite{schops2017multi} which includes  both indoor and outdoor scenes.
Recently, other large-scale stereo benchmarks dealing with driving scenarios have been released, although not yet well-established as KITTI. Among them, we mention DrivingStereo \cite{yang2019drivingstereo}, Argoverse \cite{Argoverse}, Apolloscape \cite{huang2019apolloscape} and DSEC \cite{Gehrig21ral}.
However, 
none of these more recent stereo datasets focus on the hardest open challenges for stereo matching as, instead, it is the case of our Booster dataset. Indeed, architectures ranking on top of KITTI perform remarkably well also on the above datasets.
On the contrary, we show that state-of-the-art networks struggle on Booster.

\section{Processing pipeline}
\label{sec:pipeline}
\begin{figure}
    \centering
    \begin{tabular}{cc}
        \includegraphics[height=0.18\textwidth]{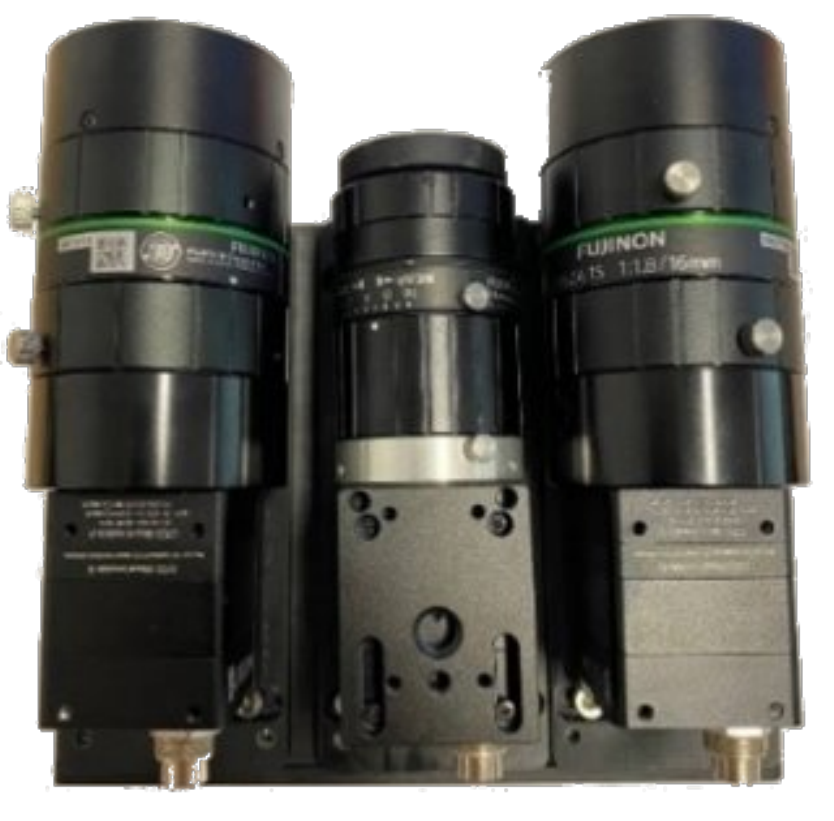}
         &
        \includegraphics[height=0.18\textwidth]{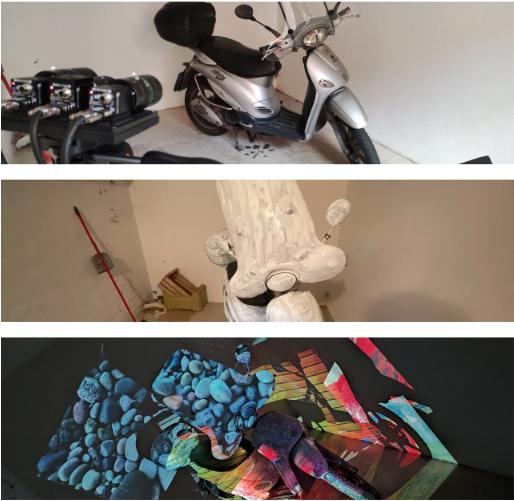} \\
    \end{tabular}
    \vspace{-0.25cm}\caption{\textbf{Cameras setup and acquisition stages.} On left, we show our camera rig, in which $L$ and $R$ are two 12 Mpx cameras, and $C$ is a wide-angle 2.3 Mpx camera. On right: i) acquisition of passive stereo pairs, ii) painting of reflective/transparent surfaces, iii) acquisition of textured stereo pairs.}
    \label{fig:rig}
\end{figure}

\textbf{Camera setup and calibration.} To collect our dataset, we have built a custom stereo rig made of 2 high resolution cameras featuring a Sony IMX253LQR-C 12.4 Mpx sensor and a lower resolution camera equipped with a Sony IMX174LQJ-C 2.3 Mpx sensor mounted between the former two, as shown in Fig. \ref{fig:rig} (left picture). From left to right we denote as $L$, $C$, and $R$ the three cameras, with $L$ providing the reference image for both the balanced ($L$,$R$) and unbalanced  ($L$,$C$) stereo pairs, and the baselines of these two setups being $\sim$ 8 and 4 centimeters, respectively.

Before acquiring the dataset, we need to calibrate our rig, in particular the two stereo systems $L-C$ and $L-R$. 
Fig. \ref{fig:pipeline} includes an overview of the calibration procedure, with a more detailed description provided in the \textbf{supplement}.

\textbf{Image acquisition.} Our trinocular rig  has been embodied into a portable setup in order to acquire a variety of scenes across different environments. In addition, our setup includes six portable projectors used to enrich the scene with random textures during the acquisition of the stereo pairs endowed  with ground-truth (red block of Fig. \ref{fig:pipeline}).
For each ground-truth acquisition, before starting, we properly setup the stage in order to capture one or more objects/surfaces embodying some of the open-challenges peculiarly addressed by our dataset.  Then, the image acquisition pipeline follows three main steps, visually resumed in Fig. \ref{fig:rig} (right pictures): i) passive images acquisition -- we collect a set of balanced and unbalances stereo images under different lightning conditions. ii) scene painting -- we carefully cover any specular/transparent surface in the scene with paint, thus allowing to properly project texture over them. iii) textured images acquisition -- we project random patterns from multiple directions and acquire a hundred images with varying textures.
Differently from the white-black banded patterns often used for this purpose \cite{davis2003spacetime}, we project color textures since we  exploit state-of-the-art deep stereo networks to label the scene. We empirically observed that color patterns result more distinctive for a deep stereo network, that is used to process bright colors typical of the synthetic datasets \cite{mayer2016large} where they have been trained. 
The outcome of this procedure consists of a set of passive stereo pairs -- both unbalanced and at high-resolution -- with different illumination conditions, representing the actual images that will be released with the dataset, and a larger set of textured images -- these latter used to produce ground-truth disparities only, as detailed in the next paragraph, and thus acquired at high-resolution only in order to produce the finest annotations. 

\textbf{Deep space-time stereo processing.} Once a set of multiple high-resolution stereo pairs -- augmented with distinctive colorful textures according to the aforementioned strategy -- has been acquired for a scene, we deploy our deep space-time stereo pipeline to infer a dense and accurate disparity map for the passive pair.
Purposely, we leverage a pre-trained deep stereo network achieving high zero-shot generalization accuracy. We expect that, in the presence of the distinctive colorful texture we project in the scene as described before, the deep network can correctly infer a reliable disparity map. Moreover, we exploit the availability of multiple stereo pairs to further improve the outcome.

Driven by the observation that most stereo networks process a \textit{cost-volume}, we accumulate all the cost volumes computed from each textured single stereo pair into an aggregated one. The resulting volume will reduce the effect of noise due to portions of the scene that may turn out not properly textured in a single acquisition. Purposely, we select RAFT-Stereo \cite{lipson2021raft}, as the top-1 method on the Middlebury 2014 stereo benchmark at the time of writing. Specifically, it uses the dot product as a measure of visual similarity between features $\mathbf{f}$ and $\mathbf{g}$ extracted respectively from the reference and target images.
Thus, RAFT-Stereo computes a correlation volume storing the inner product between any pixel features in the reference image and all those at the same y-coordinate on the target image:

\begin{equation}
    \mathbf{C}_{ijk} = \sum_h \mathbf{f}_{ijk} \cdot \mathbf{g}_{ikh}, \quad\quad \mathcal{C} \in \mathbb{R}^{H\times W\times W}
\end{equation}
Then, the network recursively estimate a disparity map $\mathbf{d}_i$ by means of a correlation look-up mechanism, implemented as a recurrent neural network $\Theta$ processing reference image features $\mathbf{f}$, some additional context features $\mathbf{c}$, the correlation volume $\mathbf{C}$ and the disparity $\mathbf{d}_{i-1}$ estimated at the previous iteration

\begin{equation}
    \mathbf{d}_i = \Theta(\mathbf{f},\mathbf{c},\mathbf{d}_{i-1},\mathbf{C}) 
\end{equation}

until a final disparity map $\mathbf{d}$ is estimated after a fixed number of iterations.
We exploit the availability of $T$ stereo pairs and build an accumulated correlation volume $\mathbf{C}^*$ by averaging the correlation volumes computed from $\mathbf{f}^t$ and $\mathbf{g}^t$ extracted from a single stereo pair $t$

\begin{equation}
    \mathbf{C}^*_{ijk} = \frac{1}{T} \sum_t \sum_h \mathbf{f}^t_{ijk} \cdot \mathbf{g}^t_{ikh}, \quad\quad \mathcal{C} \in \mathbb{R}^{H\times W\times W}
\end{equation}
Then, we exploit this enriched volume to estimate a set of disparity maps from any given stereo pair

\begin{equation}
    \mathbf{d}^t_i = \Theta(\mathbf{f}^t,\mathbf{c}^t,\mathbf{d}^t_{i-1},\mathbf{C}^*) 
\end{equation}
Once the disparity maps $\mathbf{d}^t$ have been estimated, we finally compute their average to obtain an initial, ground-truth disparity map $\mathbf{d}^*$ as well as an uncertainty guess $\mathbf{u}^*$ through their variance.
The pipeline sketched so far is effective at estimating accurate ground-truths up to half the resolution of our textured images, \ie about 6 Mpx, since RAFT-Stereo has never observed samples at such higher resolution and with such higher disparity range.
Thus, the outcome of our deep space-time stereo pipeline is a set of accurate disparity maps which yet require additional processing.

\begin{figure*}
    \centering
    \renewcommand{\tabcolsep}{10pt}
    \begin{tabular}{ccccc}
        
        \small \textit{RGB \& Mask} & \small \textit{Raft Passive} & \small \textit{Raft Spacetime} & \small \textit{SR \& Sharpening} & \small \textit{Manual Filtering}\\ 
        
        \includegraphics[width=0.12\textwidth]{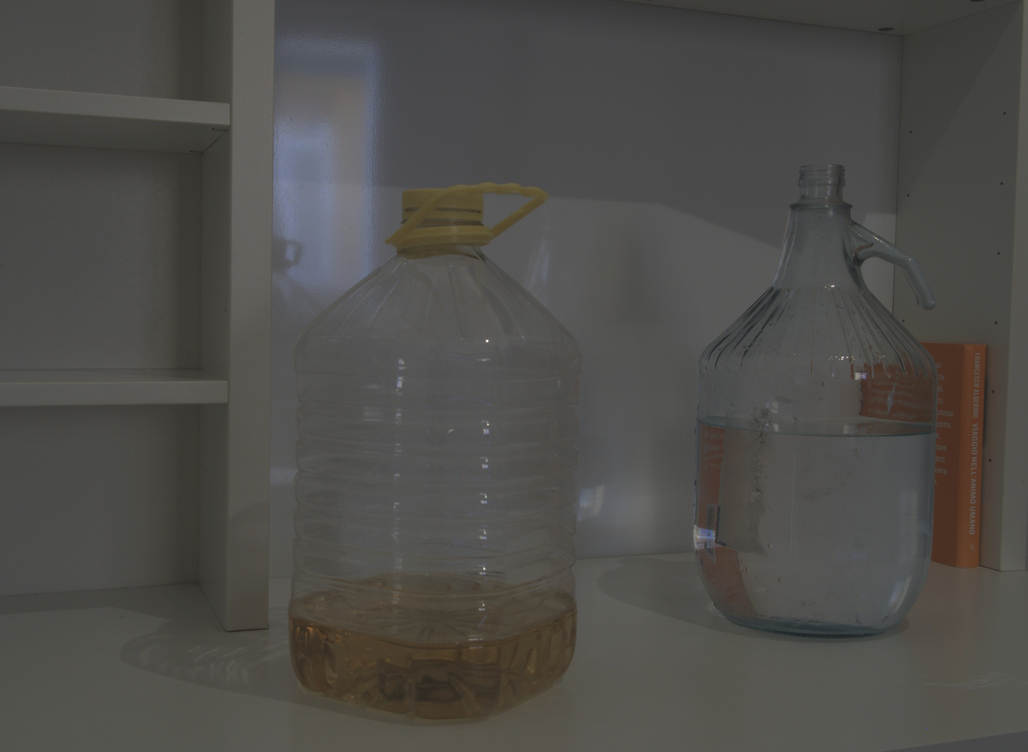} &
        \includegraphics[width=0.12\textwidth]{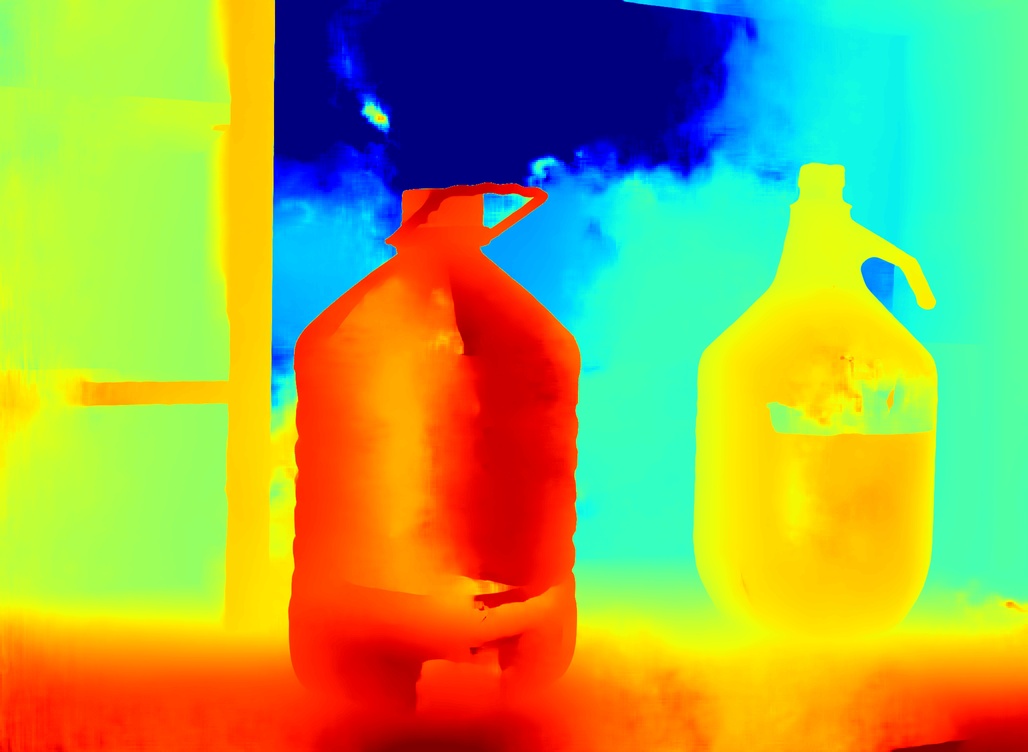} &
        \includegraphics[width=0.12\textwidth]{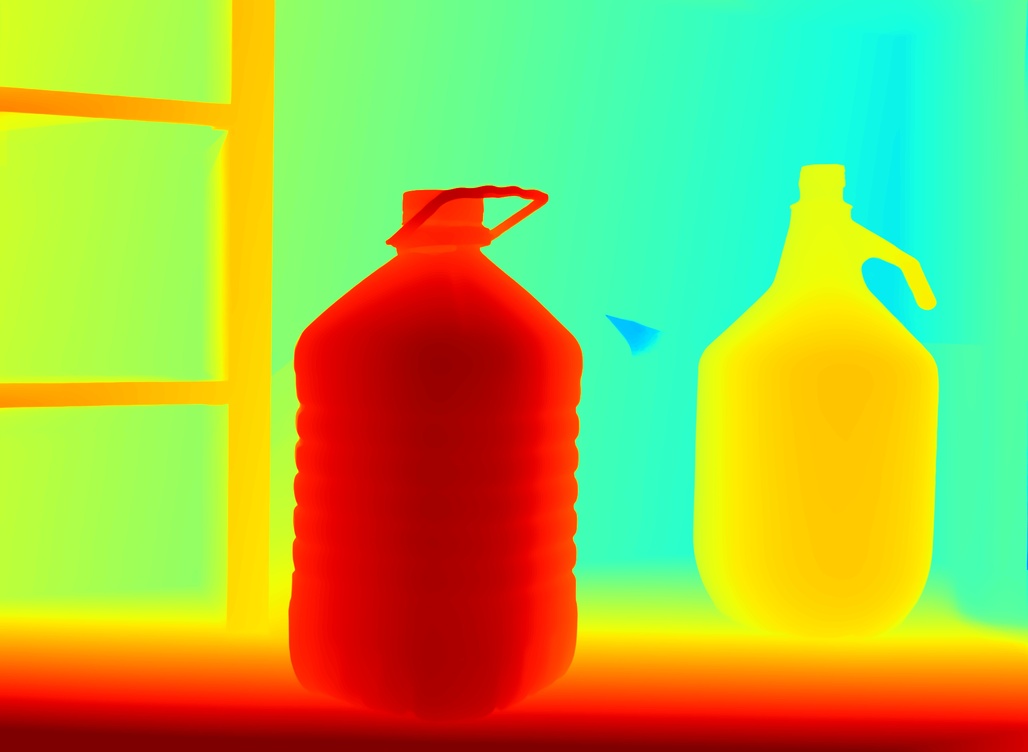} &
        \includegraphics[width=0.12\textwidth]{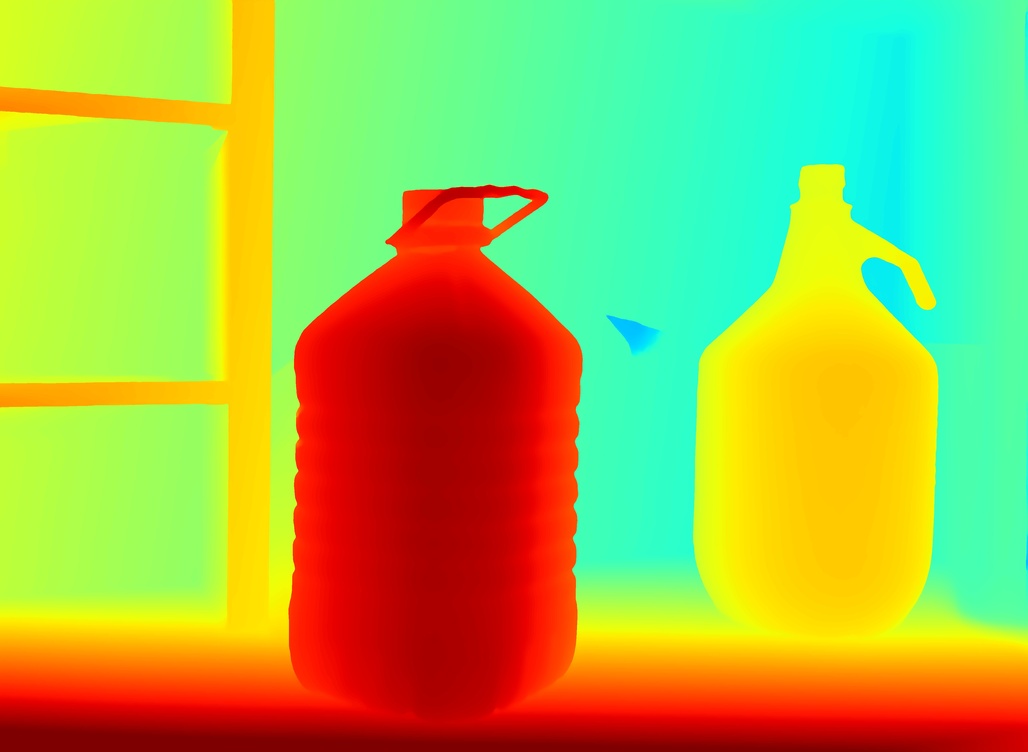} &
        \includegraphics[width=0.12\textwidth]{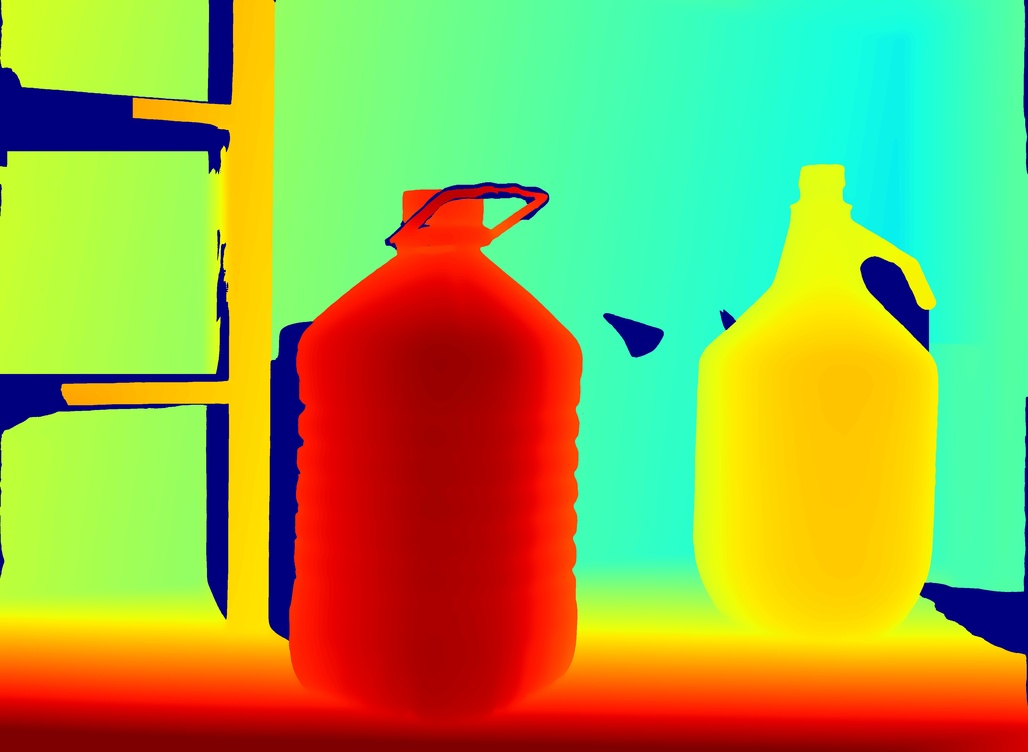} \\

        \includegraphics[width=0.12\textwidth]{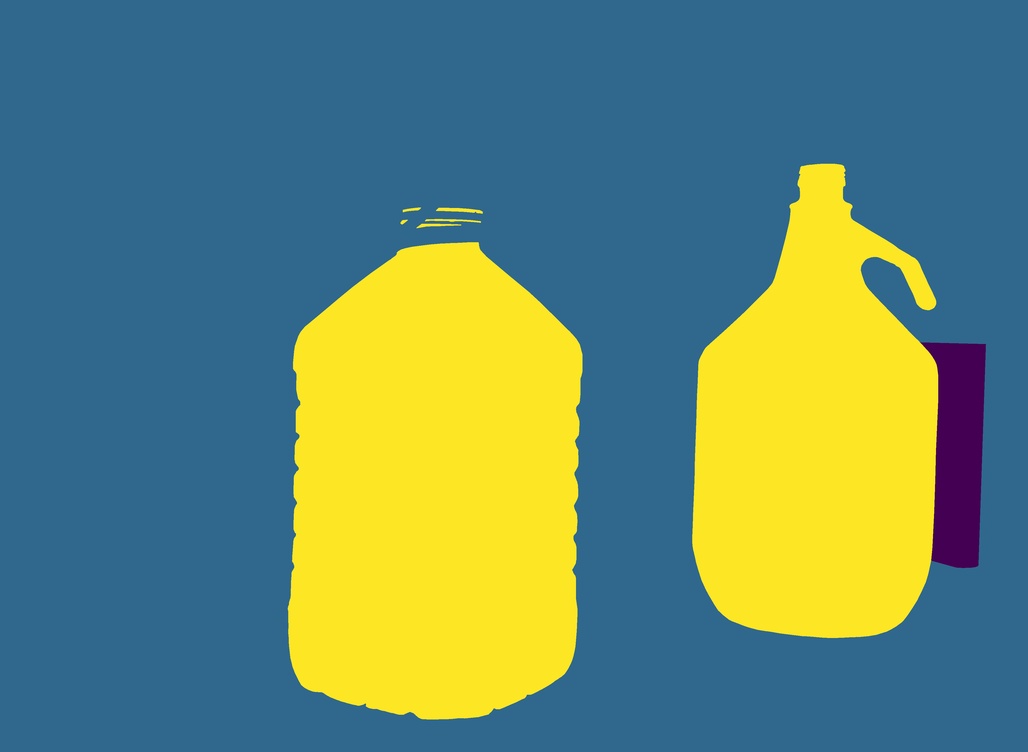} &
        \includegraphics[width=0.12\textwidth]{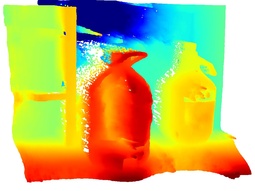} &
        \includegraphics[width=0.12\textwidth]{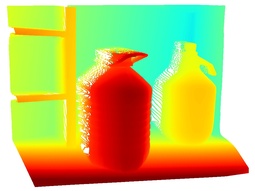} &
        \includegraphics[width=0.12\textwidth]{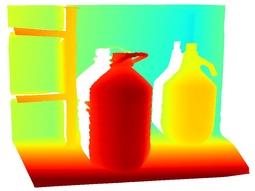} &
        \includegraphics[width=0.12\textwidth]{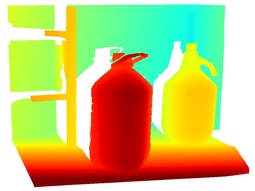} \\

    \end{tabular}
    \vspace{-0.25cm}\caption{\textbf{Data annotation pipeline.} From left to right: reference image (top) and material segmentation mask (bottom), disparity maps (top) and point clouds (bottom) obtained by RAFT-Stereo on the passive pairs, by our deep space-time stereo algorithm, by the super resolution \& sharpening procedure, and after manual cleaning.}
    \label{fig:annotation_pipeline}
\end{figure*}

\textbf{Super-resolution and sharpening.} The quality of the disparity labels produced so far is dampened by two main causes, i) the resolution, being half of the real image resolution and ii) the presence of over-smoothed depth discontinuities, a common concern in disparity maps predicted by deep networks \cite{chen2019over,Tosi2021CVPR}. To address both at once, we deploy the neural disparity refinement architecture proposed in \cite{aleotti2021neural}.  
However, being our images at a much higher-resolution compared to existing datasets, we pretrain the refinement network following \cite{aleotti2021neural}, then we overfit a single instance of it on each scene, assuming the disparity map as both input and ground-truth. This strategy allows us to preserve accurate disparity values at high-resolution while sharpening depth boundaries thanks to the network output formulation. 

Besides, we replaced the sub-pixel prediction mechanism described in \cite{aleotti2021neural} with the SMD head proposed by Tosi \etal \cite{Tosi2021CVPR}, since we empirically observed that the former introduces undesired artefacts in our setting. Thus, each neural disparity refinement network is optimized to infer a bimodal Laplacian distribution

\begin{equation}
    \mathbf{p}(d) = \frac{\pi}{2\mathbf{b}_1}e^{-\frac{\mathbf{d}^*-\mathbf{\mu}_1}{\mathbf{b}_1}} + \frac{1-\pi}{2\mathbf{b}_2}e^{-\frac{\mathbf{d}^*-\mathbf{\mu}_2}{\mathbf{b}_2}}
\end{equation}
Once the network is trained, a sharpened disparity map $\mathbf{d}^*$ is obtained at full resolution by exploiting the continuous representation enabled by the refinement network, selecting the mode with highest density value.
Concerning the implementation, a shared refinement network is pre-trained on SceneFlow following the guidelines in \cite{aleotti2021neural}. Then, a single instance is overfitted on each scene for about 300 steps before inferring the refined disparity map.

\textbf{Manual cleaning and filtering.} 
Once a full-resolution disparity map has been obtained, we manually clean it from any remaining artefact. To this aim, we project it into a 3D point cloud to better visualize structural errors in the geometry of the scene. We use the variance map $\mathbf{u}^*$ as a guidance during this operation, allowing to easily detect most of the artifacts. Points removed from the point cloud are then filtered out from the disparity map as well. 
Finally, we apply a $35\times35$ bilateral filter -- with $\sigma_\text{color}=5$ and $\sigma_\text{dist}=50$ -- to smooth objects surfaces and obtain the final map $\mathbf{d}^*$.

Fig. \ref{fig:annotation_pipeline} illustrates the pipeline described so far, showing the increasing quality of 3D reconstruction yielded by our annotations after each step.

\textbf{Accuracy assessment.} We follow the strategy used by Scharstein \etal \cite{scharstein2014high} in the Middlebury 2014 dataset to measure the accuracy of our ground-truth annotations. Accordingly, we manually select planar regions from the images and fit a plane to the recovered disparities over each of them, then we measure the residuals between the fitted plane equation and the actual disparities. We perform this evaluation over 153 planar regions achieving an average residual error of 0.053, which turns out comparable to that reported for the Middlebury 2014 dataset (0.032), yet without applying an explicit sub-pixel refinement based on plane fitting. 

\textbf{Left-right consistency (balanced setup).} We also filter out occluded pixels by performing a left-right consistency check. Purposely, the processing pipeline described so far is performed twice for each scene, producing two disparity maps, $\mathbf{d}^*_L$ and $\mathbf{d}^*_R$, for the left and right images, respectively. Then, any pixel at coordinates $(x,y)$ in $\mathbf{d}^*_L$ is filtered out in case the absolute difference with its match $x-\mathbf{d}_L(x,y),y$ in $\mathbf{d}^*_R$ is larger than a threshold, set to 2 pixels in our case

\begin{equation}
    |\mathbf{d}_L(x,y)-\mathbf{d}_R(x-\mathbf{d}_L(x,y),y)| > 2
\end{equation}
The same procedure is performed on top of $\mathbf{d}^*_R$, removing any pixel at coordinate $(x,y)$ after comparison with pixel $(x+\mathbf{d}^*_R(x,y),y)$ on the left disparity map. 

The output of our overall annotation pipeline consists of three high-resolution ground-truth disparity maps per scene: two for the left and right images of the  balanced setup, one for the unbalanced setup.

\textbf{Segmentation masks.} Finally, we manually label images to annotate challenging surfaces, \ie transparent or specular, with segmentation masks. We cluster object surfaces into 4 classes (from 0 to 3) with increasing level of transparency and/or specularity, with class 0 identifying very opaque materials (\eg, a wood table) and class 3 those highly transparent/specular (\eg, window glasses/mirrors). An example of segmentation mask is shown in Fig. \ref{fig:annotation_pipeline}.

\textbf{Warping (unbalanced setup).}
The ground-truths obtained so far are aligned with images of $L-R$. 
However, we want ground-truths also for the unbalanced $L-C$ stereo system. 
Being the rectification transformation an homography (i.e., only a change of intrinsic parameters and a rotation), we can easily perform a backward warping of the ground-truths 
of the left images of $L-R$ to align them to the left images of $L-C$. When warping disparity maps, we take into account the rotation of the camera reference frame and the different baselines of the two stereo systems before performing the warping. Additional details about the warping procedure can be found in the \textbf{supplement}. 
\section{The Booster Dataset}

\begin{figure}
    \centering
    \renewcommand{\tabcolsep}{1pt}
    \begin{tabular}{cccccc}
    \multicolumn{2}{c}{\scriptsize  \textit{Balanced Setup}} & & \scriptsize  \textit{Unbalanced Setup} & & \scriptsize  \textit{Illuminations} \\
    \includegraphics[width=0.10\textwidth]{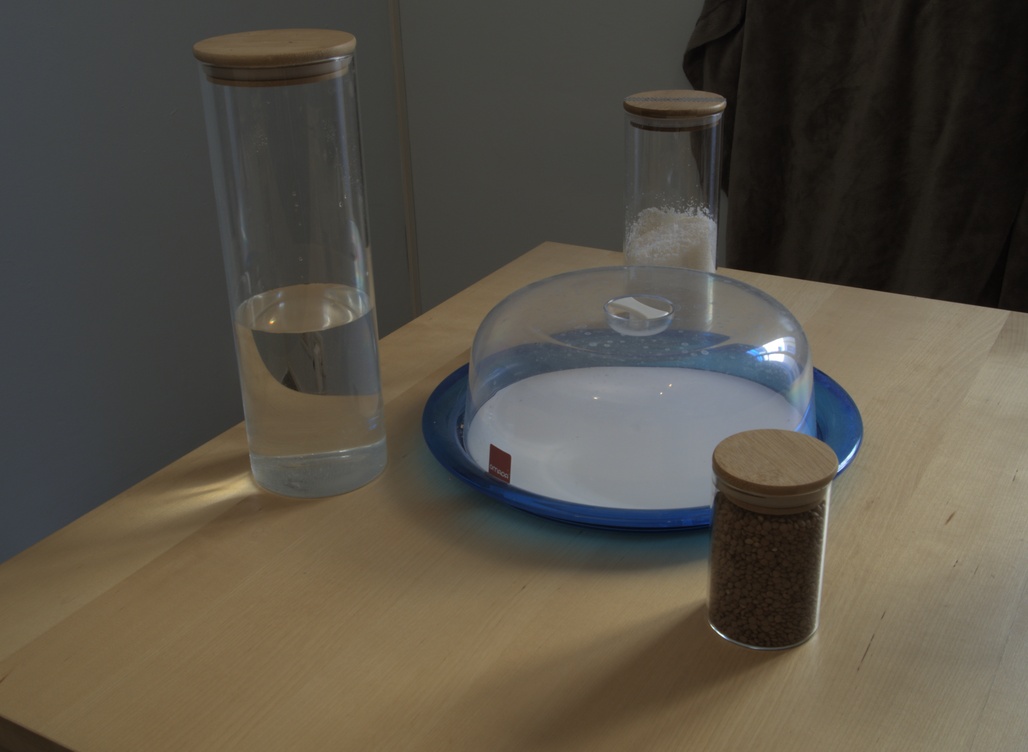} &
    \includegraphics[width=0.10\textwidth]{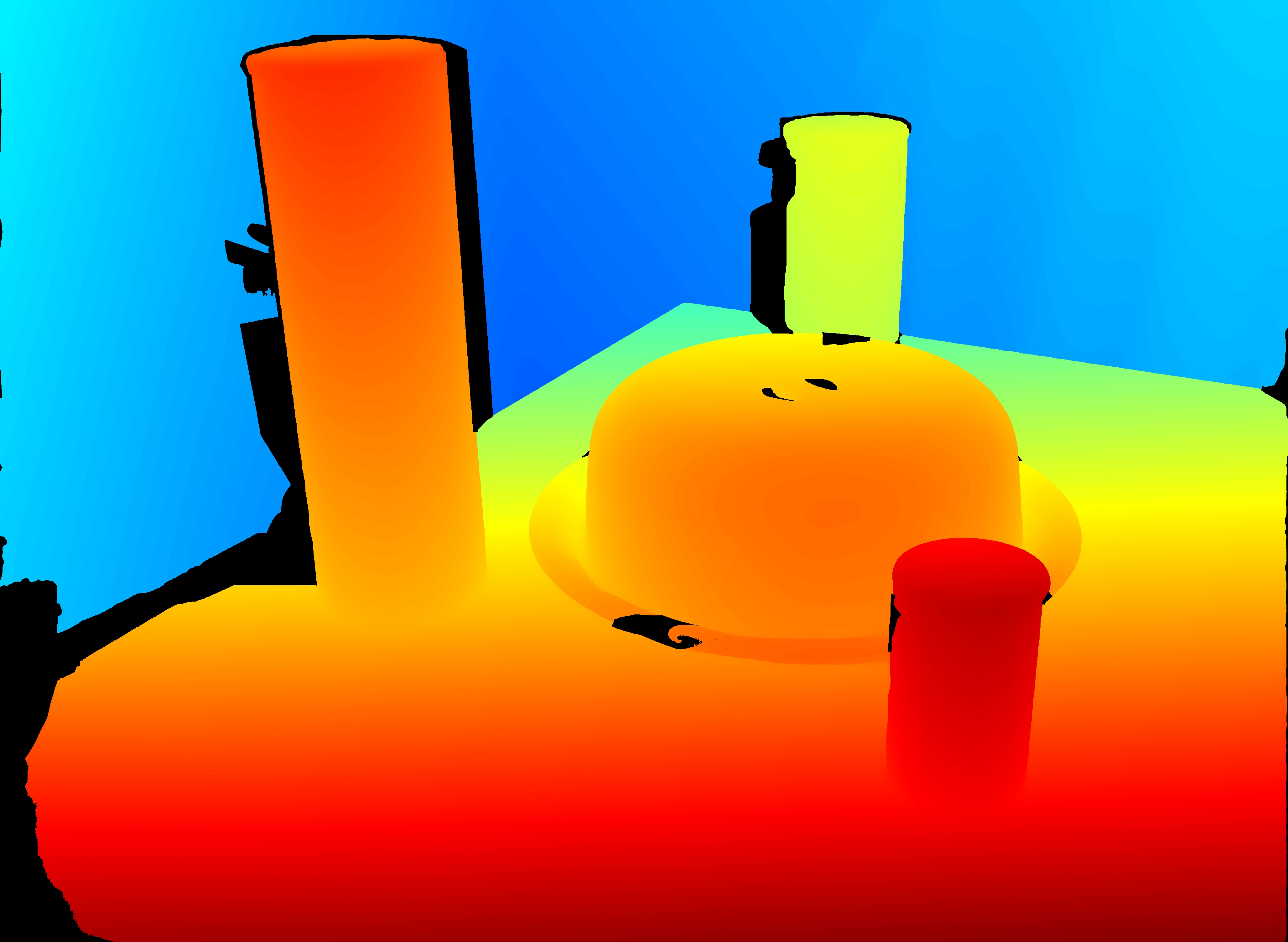} & &
    \includegraphics[width=0.10\textwidth]{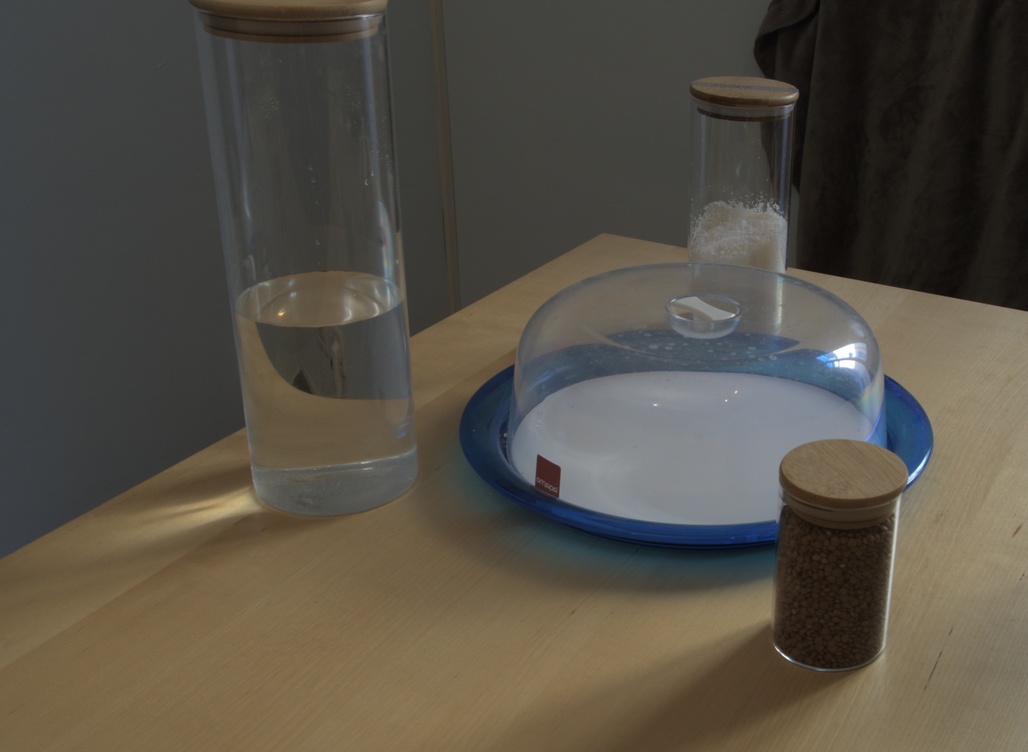} & &
    \includegraphics[width=0.10\textwidth]{images/dataset_sample/camera_00_02/im0.jpg} \\
    
    \includegraphics[width=0.10\textwidth]{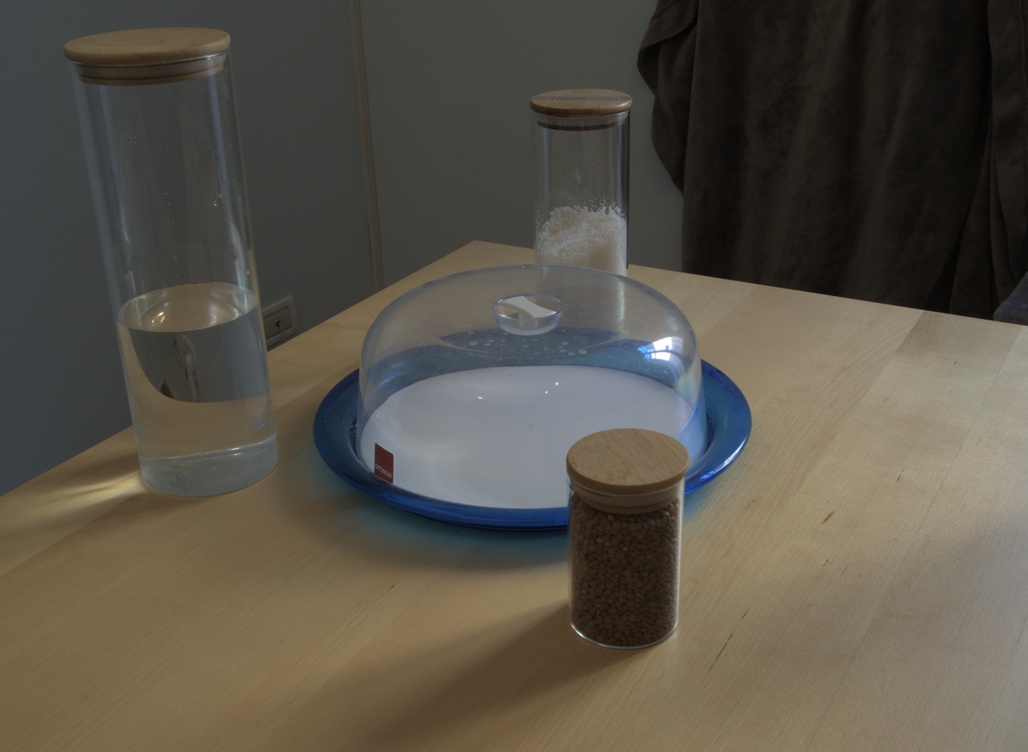} & 
    \includegraphics[width=0.10\textwidth]{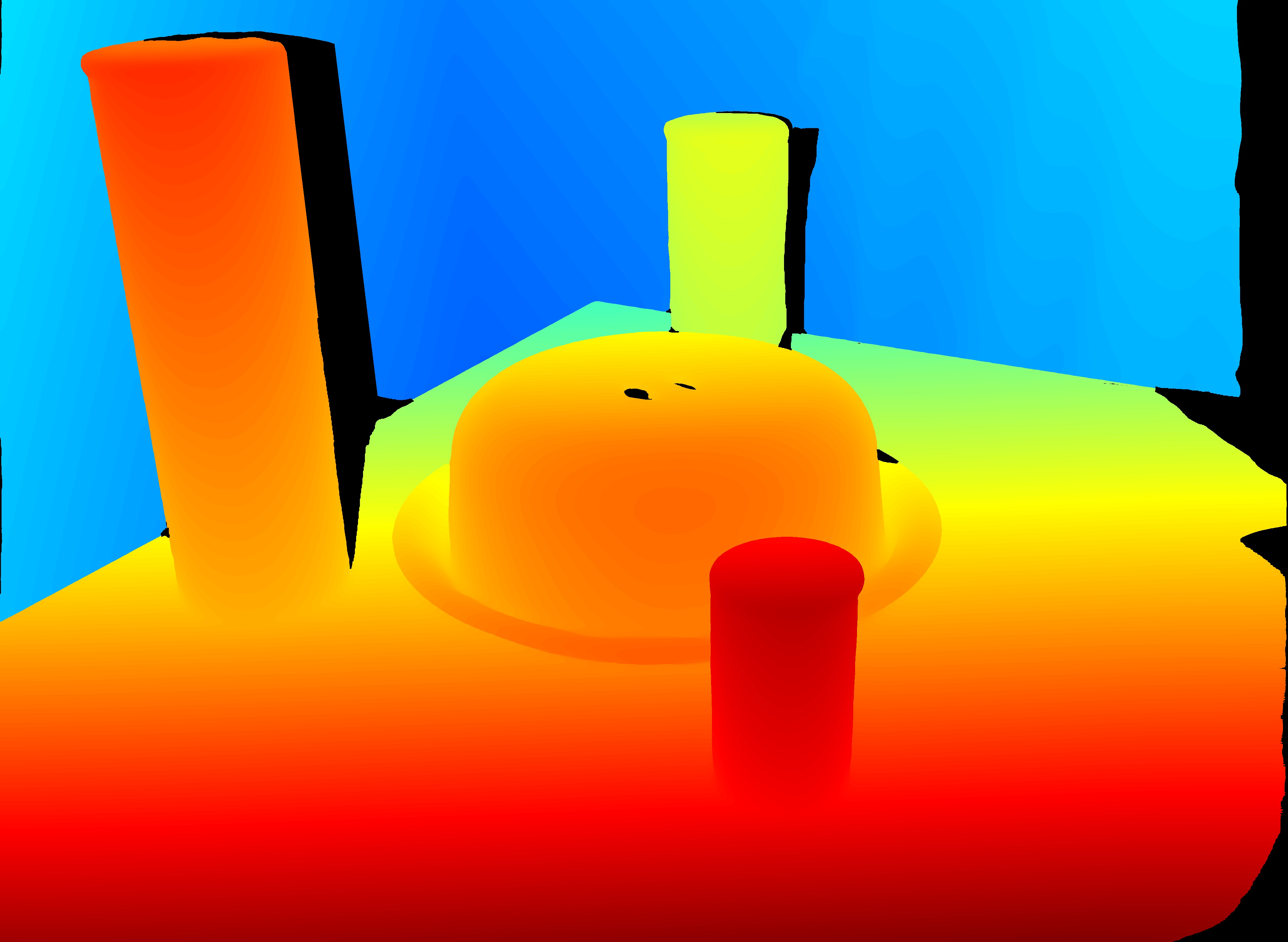} & &
    \includegraphics[width=0.04\textwidth]{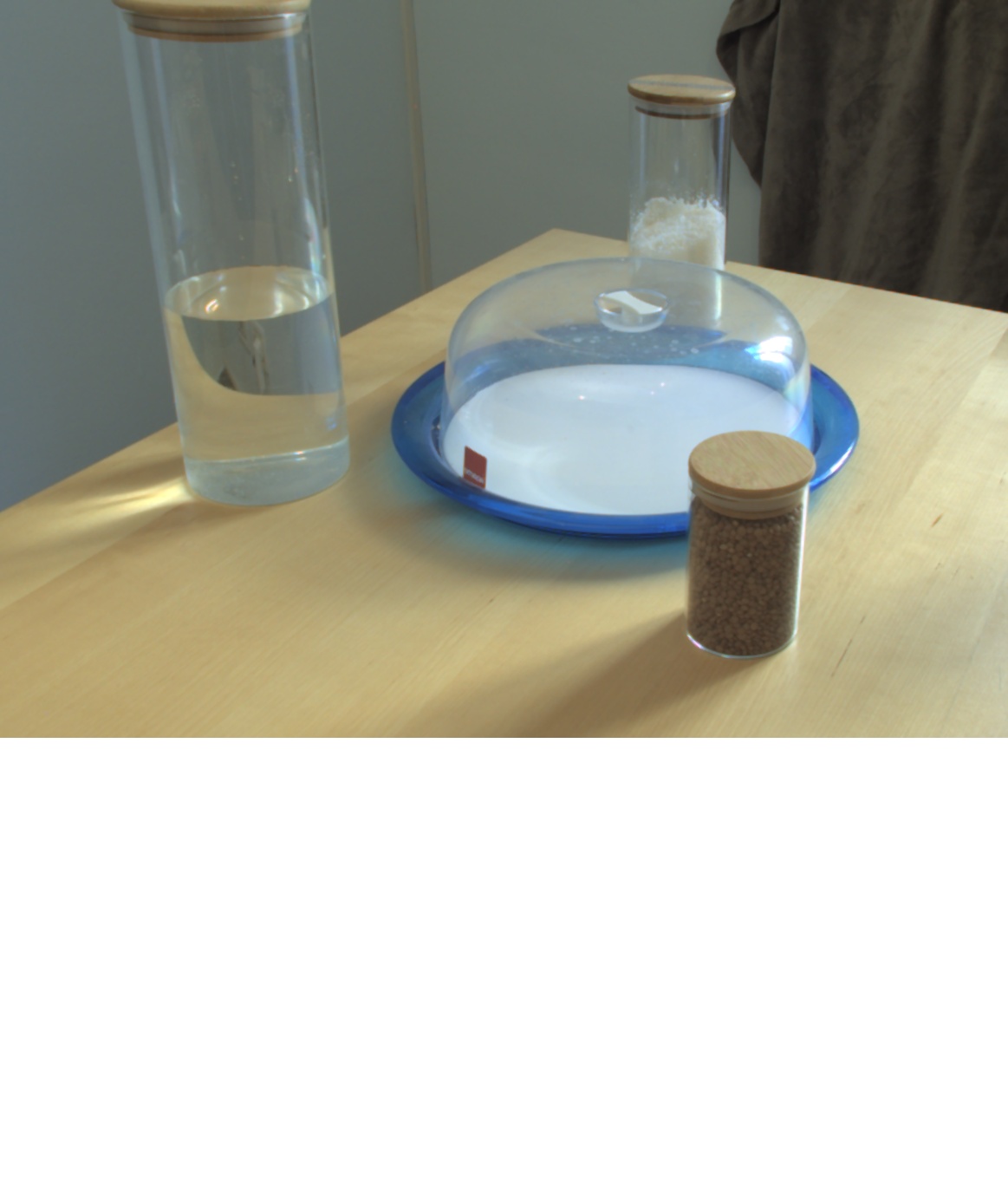} & &
    \includegraphics[width=0.10\textwidth]{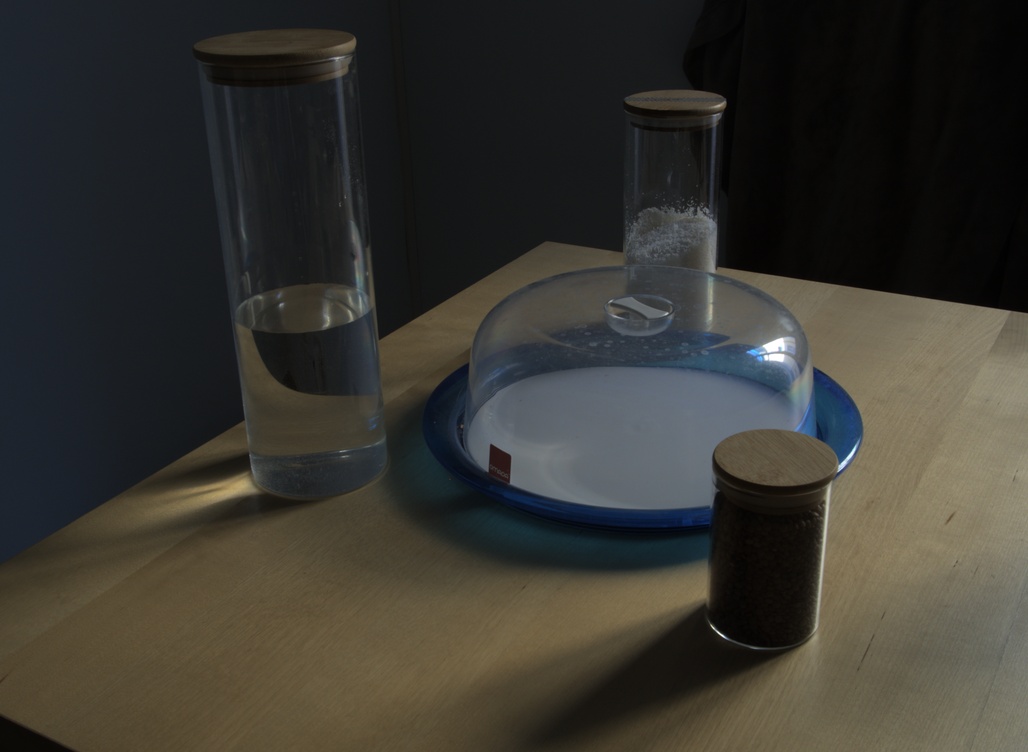} \\
    
    \includegraphics[width=0.10\textwidth]{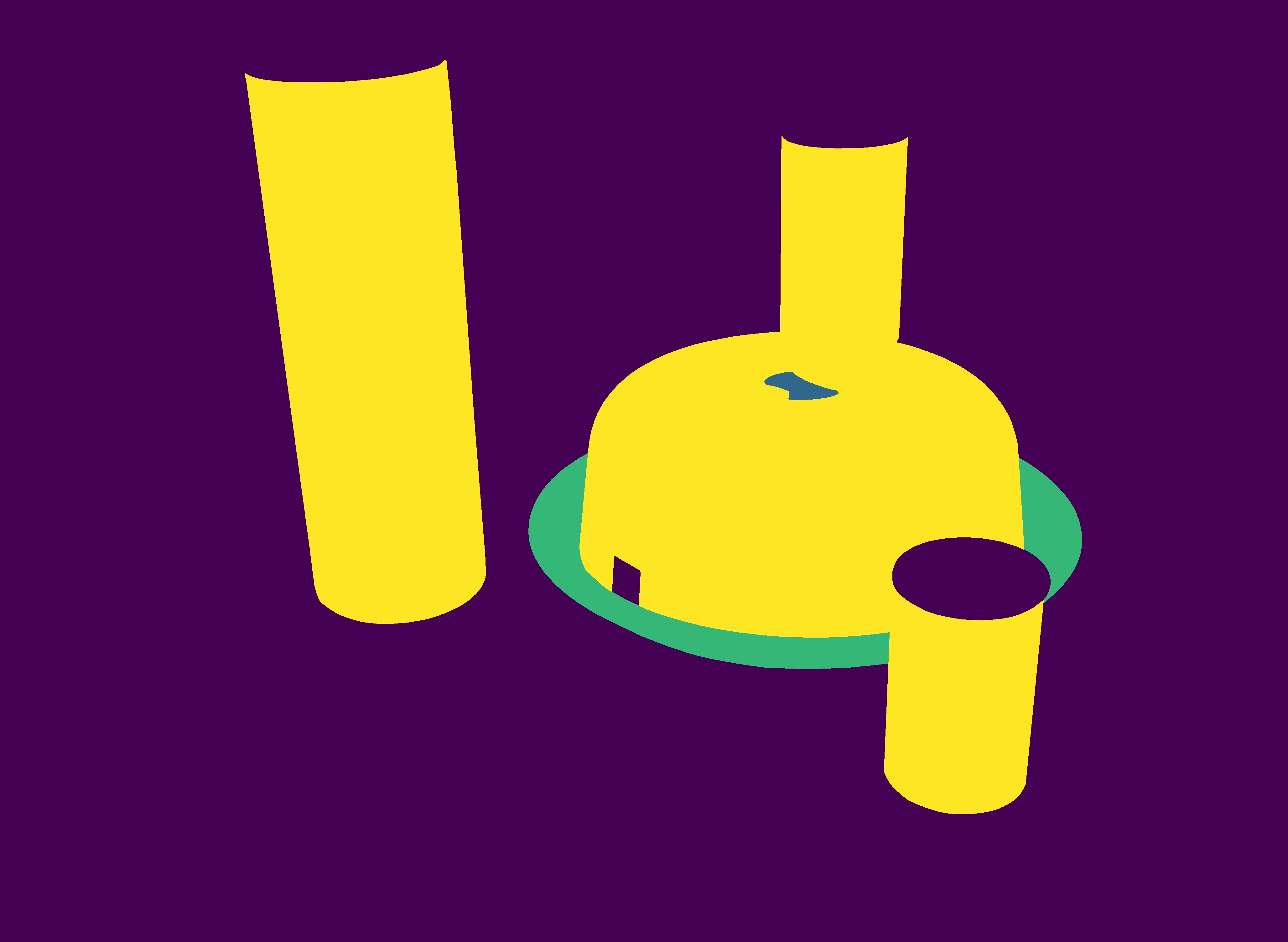} &
    \includegraphics[width=0.10\textwidth]{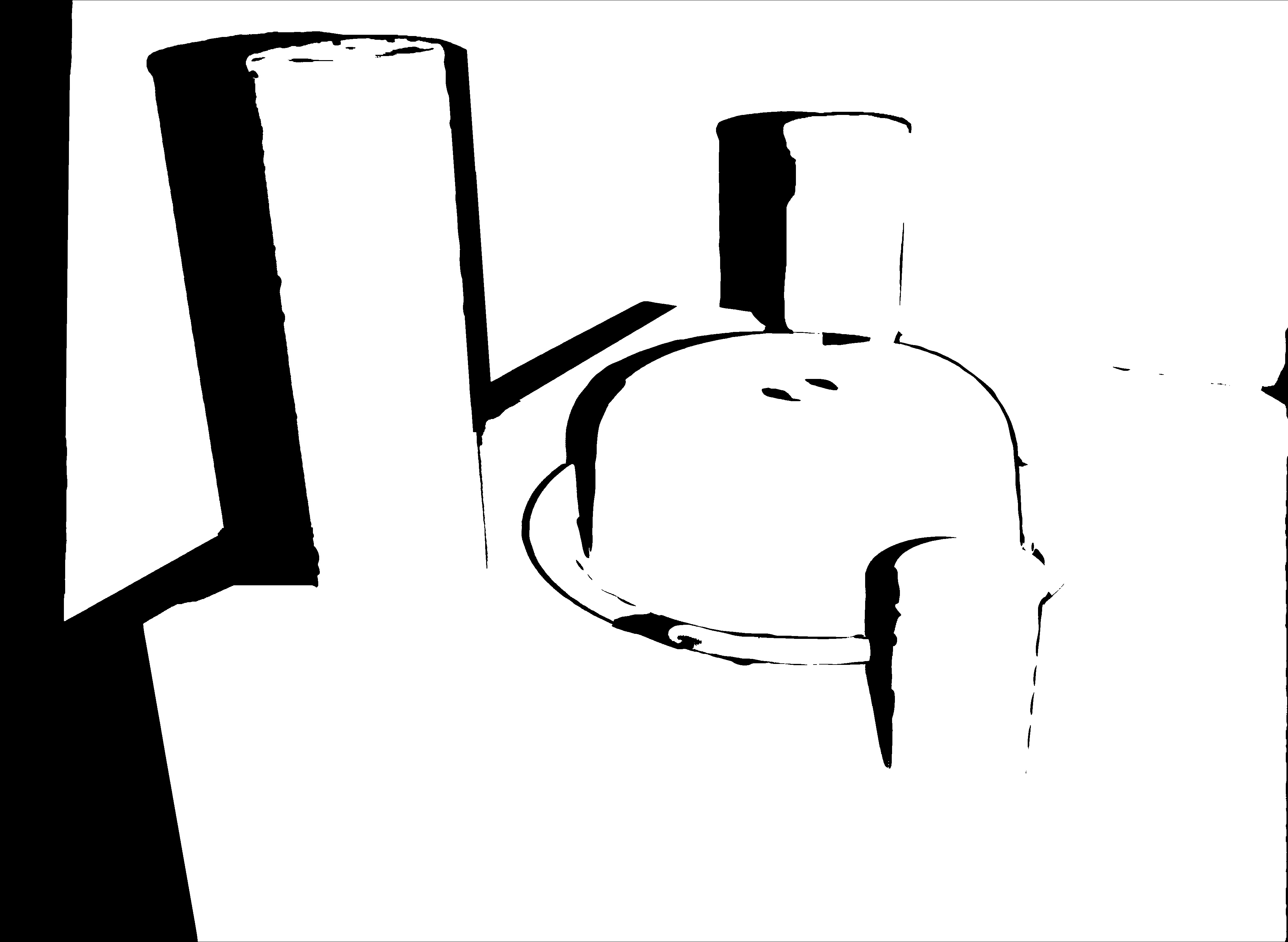} & &
    \includegraphics[width=0.10\textwidth]{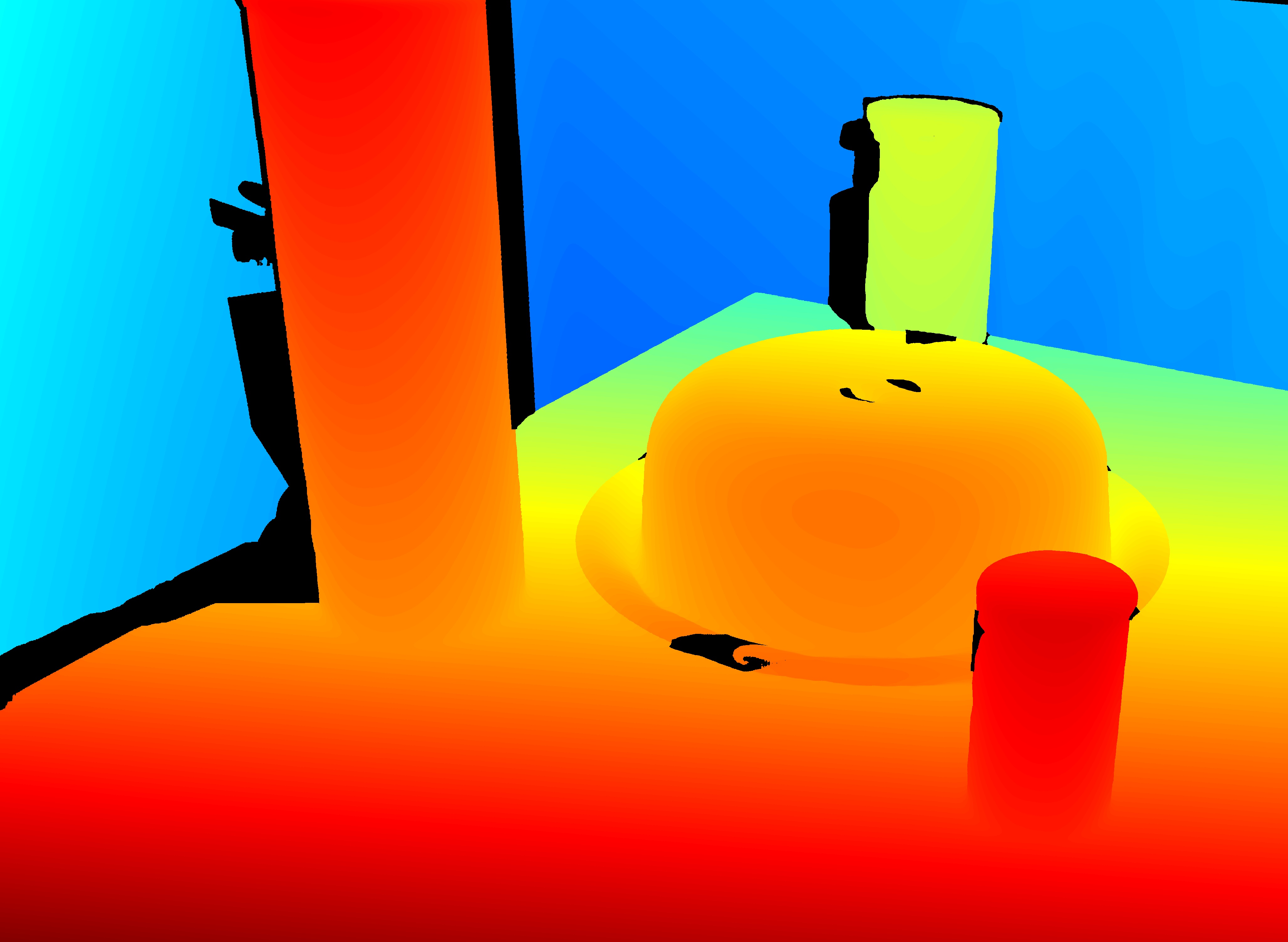} & &
    \includegraphics[width=0.10\textwidth]{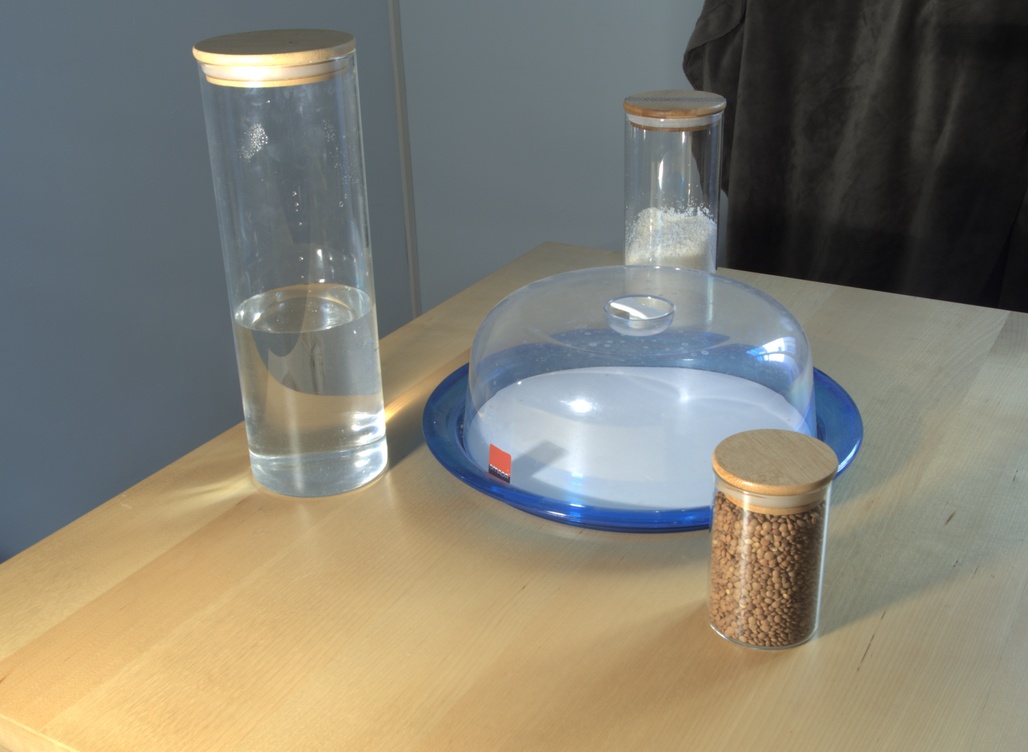} \\
    
    \end{tabular}
    \vspace{-0.25cm}\caption{\textbf{A scene  from the Booster testing split.} First two columns: data made available in the balanced setup (12 Mpx stereo pair, material segmentation mask, left and right disparity maps and left-right consistency mask). Third column: data dealing with the unbalanced setup (12 Mpx -  1.1 Mpx  image pair, high-res  disparity map associated with the  12 Mpx  image ). Last columns: additional  12 Mpx images acquired under different illuminations.}
    \label{fig:booster_sample}
\end{figure}

\textbf{Composition.} To build up the dataset, we set the stage in 64 different indoor scenes. Then, we collected a variety of passive stereo images under different illumination conditions, leading to a total of 419 stereo samples for which we obtain dense annotations through the pipeline detailed in Sec. \ref{sec:pipeline}. We split the 64 scenes into 38 and 26 for training and testing purposes, respectively. As a result, Booster counts  228 training images and 191 testing images. In defining the split we aimed at having diversity of environments between the training and test scenes as well as at achieving a balanced distribution of challenging objects and materials (\eg, both splits contain a scene framing a mirror).  Two main benchmarks are defined in Booster: the \textbf{Balanced} benchmark, including 419 stereo pairs at 12 Mpx, and the \textbf{Unbalanced} one, featuring equally many 12 Mpx - 1.1 Mpx pairs. The latter represents the first-ever real dataset for unbalanced stereo matching, a task studied so far only by simulating the unbalanced setup by resizing one of the two images of a balanced pair  same-resolution stereo images \cite{liu2020visually,aleotti2021neural}.
More details regarding dataset images are reported in the \textbf{supplement}.
Fig. \ref{fig:booster_sample} concerns a sample from the testing split and shows  the data made available for any acquired scene.  

\textbf{Unlabeled samples.} To encourage research on weakly-supervised approaches, \ie not requiring ground-truth labels at training time, we release 15K additional samples  collected --in both balanced and unbalanced settings -- in a variety of indoor and outdoor environments. 

\textbf{Evaluation metrics.} To assess the accuracy of stereo algorithms and networks, we adopt a set of metrics inspired by Middlebury 2014 \cite{scharstein2014high}. Specifically, we compute the amount of pixels having error larger than a threshold $\tau$ (bad-$\tau$). As initially our ground-truth maps are inferred at half the input resolution, we assume 2 pixels as the lowest threshold. Then, given the much higher resolution of our images, we compute error rates up to bad-8. We also measure Mean Absolute Error (MAE) and Root Mean Squared Error (RMSE). All metrics are computed on any valid pixel (\textit{All}), or in alternative, on pixels belonging to the material class $i$ (\textit{Class $i$}) to evaluate the impact of non-Lambertian objects. In the case of the balanced setup, we also evaluate on not-occluded pixels identified by the left-right check of our annotation pipeline (\eg, the bottom image in the third column of Fig. \ref{fig:booster_sample}).

\section{Experiments}

\begin{table*}[t]
\centering
\renewcommand{\tabcolsep}{12pt}
\scalebox{0.47}{
\begin{tabular}{ccc}

 \begin{tabular}{c}
    \multirow{3}{*}{\rotatebox{90}{Full res.}} \\ 
 \end{tabular}
 \begin{tabular}{ll}
 \\
 \toprule
 & Input \\
 Model & Res.\\
 \midrule
 SGM \cite{hirschmuller2007stereo} & Q \\
 MC-CNN \cite{zbontar2016stereo} & Q \\
 LEAStereo \cite{cheng2020hierarchical} & Q \\
 CFNet \cite{shen2021cfnet} & Q \\
 HSMNet \cite{yang2019hierarchical} & Q \\
 RAFT-Stereo \cite{lipson2021raft} & Q \\
 \midrule
 SGM \cite{hirschmuller2007stereo} & H \\
 HSMNet \cite{yang2019hierarchical} & H \\
 SGM+Neural Ref. \cite{aleotti2021neural} & H \\
 RAFT-Stereo \cite{lipson2021raft} & H \\
 \midrule
 HSMNet \cite{yang2019hierarchical} & F \\
 \bottomrule
 \end{tabular}
 &
 \begin{tabular}{rrrr | rr }
 \multicolumn{6}{c}{All pixels} \\
 \toprule
 bad-2 & bad-4 & bad-6 & bad-8 & MAE & RMSE \\
 (\%) & (\%) & (\%) & (\%) & (px.) &  (px.) \\
 \midrule
 80.35 & 66.89 & 58.09 & 52.21 & 57.01 & 119.21 \\ 
 88.09 & 66.30 & 47.77 & 40.53 & 31.23 & 62.98 \\ 
 70.86 & 55.41 & 47.56 & 42.25 & 27.61 & 51.72 \\ 
 61.34 & 48.33 & 42.22 & 38.34 & 27.60 & 51.62 \\ 
 66.95 & 48.05 & 37.46 & 31.14 & 20.97 & 42.72 \\ 
 \bfseries 40.27 & \bfseries 27.54 & \bfseries 22.83 & \bfseries 20.13 & \bfseries 17.08 & \bfseries 36.30 \\ 
 \midrule
 76.61 & 64.72 & 58.34 & 54.37 & 71.68 & 133.35 \\ 
 53.75 & 36.47 & 28.71 & 24.50 & 19.17 & 42.00 \\ 
 78.54 & 63.20 & 53.77 & 46.87 & 31.82 & 67.02 \\ 
 46.31 & 35.49 & 30.98 & 28.15 & 23.95 & 49.94 \\ 
 \midrule
 50.85 & 36.53 & 30.77 & 27.56 & 30.82 & 68.97 \\ 
 \bottomrule
 \end{tabular}
 &
 \begin{tabular}{rrrr | rr }
 \multicolumn{6}{c}{Cons pixels} \\
 \toprule
 bad-2 & bad-4 & bad-6 & bad-8 & MAE & RMSE \\
 (\%) & (\%) & (\%) & (\%) & (px.) &  (px.) \\
 \midrule
 78.40 & 63.70 & 54.13 & 47.79 & 41.28 & 91.86 \\ 
 87.64 & 64.20 & 44.24 & 36.70 & 27.56 & 57.34 \\ 
 69.15 & 53.17 & 45.42 & 40.24 & 26.36 & 49.52 \\ 
 59.13 & 46.02 & 40.08 & 36.36 & 25.72 & 48.55 \\ 
 65.23 & 45.86 & 35.36 & 29.31 & 20.93 & 42.42 \\ 
 \bfseries 38.65 & \bfseries 26.49 & \bfseries 22.25 & \bfseries 19.84 & \bfseries 17.13 & \bfseries 35.76\\ 
 \midrule
 74.18 & 61.17 & 54.25 & 49.99 & 55.25 & 106.55 \\ 
 51.25 & 34.06 & 26.78 & 23.01 & 18.92 & 41.28 \\ 
 78.35 & 60.59 & 49.59 & 42.50 & 30.92 & 68.37 \\ 
 44.02 & 33.59 & 29.49 & 26.95 & 23.25 & 48.11 \\ 
 \midrule
 48.11 & 33.88 & 28.50 & 25.61 & 30.02 & 66.79 \\ 
 \bottomrule
 \end{tabular}
 
 \\
 \\
 
 \begin{tabular}{c}
    \multirow{3}{*}{\rotatebox{90}{Quarter res.}} \\ 
 \end{tabular}
 \begin{tabular}{ll}
 \\
 \toprule
 & Input \\
 Model & Res.\\
 \midrule
 SGM \cite{hirschmuller2007stereo} & Q \\
 MC-CNN \cite{zbontar2016stereo} & Q \\
 LEAStereo \cite{cheng2020hierarchical} & Q \\
 CFNet \cite{shen2021cfnet} & Q \\
 HSMNet \cite{yang2019hierarchical} & Q \\
 RAFT-Stereo \cite{lipson2021raft} & Q \\
 \bottomrule
 \end{tabular}
 &
 \begin{tabular}{rrrr | rr }
 \multicolumn{6}{c}{All pixels} \\
 \toprule
 bad-2 & bad-4 & bad-6 & bad-8 & MAE & RMSE \\
 (\%) & (\%) & (\%) & (\%) & (px.) &  (px.) \\
 \midrule
 52.76 & 39.43 & 33.11 & 29.26 & 14.64 & 30.68 \\ 
 40.33 & 30.36 & 25.64 & 22.25 & 7.82 & 15.85 \\ 
 42.21 & 30.23 & 24.37 & 20.43 & 6.89 & 12.92 \\ 
 38.31 & 29.53 & 24.70 & 21.34 & 6.89 & 12.89 \\ 
 31.11 & 20.25 & 15.92 & 13.23 & 5.24 & 10.67 \\ 
 \bfseries 20.13 & \bfseries 15.13 & \bfseries 12.85 & \bfseries 11.05 & \bfseries 4.27 & \bfseries 9.05 \\ 
 \bottomrule
 \end{tabular}
 &
 \begin{tabular}{rrrr | rr }
 \multicolumn{6}{c}{Cons pixels} \\
 \toprule
 bad-2 & bad-4 & bad-6 & bad-8 & MAE & RMSE \\
 (\%) & (\%) & (\%) & (\%) & (px.) &  (px.) \\
 \midrule
 48.42 & 34.18 & 27.58 & 23.63 & 10.75 & 24.05 \\ 
 36.50 & 26.50 & 21.84 & 18.79 & 6.90 & 14.43 \\ 
 40.19 & 28.68 & 23.21 & 19.50 & 6.58 & 12.36 \\ 
 36.32 & 27.85 & 23.24 & 20.05 & 6.42 & 12.11 \\ 
 29.25 & 19.47 & 15.70 & 13.23 & 5.22 & 10.59 \\ 
 \bfseries 19.82 & \bfseries 15.19 & \bfseries 12.98 & \bfseries 11.17 & \bfseries 4.28 & \bfseries 8.91 \\ 
 \bottomrule
 \end{tabular}
 
 \end{tabular}
 }
 \vspace{-0.25cm}\caption{\textbf{Results on the Booster Balanced testing split.} We run off-the-shelves stereo networks, using weights provided by their authors. We evaluate on full resolution ground-truth maps, or by downsampling them to quarter resolution. Best scores in \textbf{bold}.}
 \label{tab:stereo_tournament}
\end{table*}

\begin{table}[t]
\centering
\scalebox{0.6}{
\begin{tabular}{ccc}

 \begin{tabular}{c}
    \multirow{3}{*}{\rotatebox{90}{Full res.}} \\ 
 \end{tabular}
 \begin{tabular}{l}
 \\
 \toprule
 \\
 Category \\
 \midrule
 All \\
 \midrule
 Class 0 \\
 Class 1 \\
 Class 2 \\
 Class 3 \\
 \bottomrule
 \end{tabular}
 &
 \begin{tabular}{rrrr | rr }
 \multicolumn{6}{c}{All pixels} \\
 \toprule
 bad-2 & bad-4 & bad-6 & bad-8 & MAE & RMSE \\
 (\%) & (\%) & (\%) & (\%) & (px.) &  (px.) \\
 \midrule
 40.27 & 27.54 & 22.83 & 20.13 & 17.08 & 36.30 \\
 \midrule
 32.81 & 16.67 & 11.11 & 7.92 & 3.72 & 9.38 \\
 42.95 & 27.47 & 21.60 & 18.21 & 10.20 & 19.96 \\
 73.59 & 60.69 & 51.03 & 44.51 & 36.67 & 47.44 \\
 81.54 & 71.93 & 65.22 & 59.62 & 47.73 & 59.38 \\ 
 \bottomrule
 \end{tabular}
 
 \\
 \\

 \begin{tabular}{c}
    \multirow{3}{*}{\rotatebox{90}{Quarter res.}} \\ 
 \end{tabular}
 \begin{tabular}{l}
 \\
 \toprule
 \\
 Category \\
 \midrule
 All \\
 \midrule
 Class 0 \\
 Class 1 \\
 Class 2 \\
 Class 3 \\
 \bottomrule
 \end{tabular}
 &
 \begin{tabular}{rrrr | rr }
 \multicolumn{6}{c}{All pixels} \\
 \toprule
 bad-2 & bad-4 & bad-6 & bad-8 & MAE & RMSE \\
 (\%) & (\%) & (\%) & (\%) & (px.) &  (px.) \\
 \midrule
 20.13 & 15.13 & 12.85 & 11.05 & 4.27 & 9.05 \\
 \midrule
 7.97 & 3.72 & 2.33 & 1.82 & 0.93 & 2.28 \\
 18.22 & 11.22 & 7.84 & 6.68 & 2.55 & 4.97 \\
 44.47 & 32.14 & 27.92 & 25.43 & 9.17 & 11.88 \\
 59.65 & 43.99 & 34.97 & 28.51 & 11.92 & 14.82 \\ 
 \bottomrule
 \end{tabular} \\
 
 \end{tabular}
 }
 \vspace{-0.25cm}\caption{\textbf{Results on the Booster Balanced testing split -- material segmentation.} We run RAFT-Stereo \cite{lipson2021raft}, using weights made available by their authors and process quarter resolution images. We evaluate on full resolution ground-truth maps, or by downsampling them to quarter resolution.}
 \label{tab:segmentation}
\end{table}

\subsection{Balanced Stereo Benchmark} 

We start by considering the Balanced split of Booster and perform a set of different experiments. 

\textbf{Off-the-shelf deep networks.} We run a set of off-the-shelf, state-of-the-art deep stereo networks on the test set of Booster in order to assess their accuracy.  We select networks with freely available implementations and pretrained weights providing  good performance on the Middlebury 2014 dataset, \ie the most challenging among existing benchmarks. This constraint limits our selection to HSMNet \cite{yang2019hierarchical} LEAStereo \cite{cheng2020hierarchical}, CFNet \cite{shen2021cfnet}, RAFT-Stereo \cite{lipson2021raft} and Neural Disparity Refinement \cite{aleotti2021neural}. 
We also evaluate, as references, the popular Semi-Global Matching algorithm (SGM) \cite{hirschmuller2007stereo} and the pivotal MC-CNN network \cite{zbontar2016stereo} in its fast variant because of memory constraints. 

Tab. \ref{tab:stereo_tournament} collects the outcome of this evaluation. In the top portion of the table, we compare the predicted disparity maps with full-resolution ground-truths, on \textit{All} (left) and \textit{Cons} (right) pixels, the latter being the pixels of the left image that turn out consistent upon performing the left-right check and, as such, are considered as not-occluded. Each method processes input images either at the original resolution (F) or scaled to half (H) or quarter (Q) resolution. Deep networks inferences are performed on a single 3090 RTX GPU. We can notice how most methods can run only at Q resolution, mainly because of memory constraints. Consequently, their output is upsampled with nearest-neighbor interpolation in order to perform the comparison with the full-resolution ground-truth maps, with disparities scaled by the upsampling factor itself. We can notice how all methods struggle at achieving good results at such high resolution, with RAFT-Stereo achieving the best results -- not surprisingly, perhaps, given its top-rank on Middlebury. Error metrics computed on \textit{All} and \textit{Cons} pixels yield similar scores, proving that occlusions do not represent the main difficulties in our benchmark.
In the bottom portion of Tab. \ref{tab:stereo_tournament}, predicted disparities are compared with ground-truth disparity maps downsampled to a quarter (Q) of the original resolution. Although the error metrics are much lower in general, we point out how they are still very far from those observed on existing benchmarks \cite{Geiger2012CVPR,Menze2015CVPR,scharstein2014high,schoeps2017cvpr}, confirming that resolution is certainly a challenge in our benchmark, yet not the only one -- due to the large presence of transparent and specular surfaces framed during acquisition.

\textbf{Evaluation on challenging regions.} We dig deeper into the unique features of Booster by evaluating the accuracy of the predicted disparities in regions of increasing level of difficulty, as defined by means of the material segmentation masks. Purposely, we select the top-performing network from the previous evaluation, \ie RAFT-Stereo, and evaluate it on subsets of pixels defined by our manually annotated masks. Tab. \ref{tab:segmentation} collects the outcome of this evaluation, together with results on all valid pixels as a reference. Starting from the least challenging category, we observe much lower error scores -- in particular, by evaluating on quarter resolution ground-truths (bottom), we achieve results comparable with those of existing benchmarks \cite{scharstein2014high}. By gradually increasing the degree of difficulty of the considered pixels, we witness a large increase of the errors. This confirms both our claims on the open-challenges in deep stereo as well as the significance of our segmentation masks.

\textbf{Fine-tuning by the Booster training data.} Finally, we fine-tune RAFT-Stereo on the Booster training set to show that the availability of annotated scenes can be effective in improving the result in presence of the open challenges addressed in this paper. We run 100 epochs on batches of two 884$\times$456 crops, extracted from images randomly resized to half or quarter of the original resolution, using the optimization procedure from \cite{lipson2021raft} and initial learning rate set to 1e-5.

Tab. \ref{tab:finetuning} collects the results on \textit{All} pixels, as well as on each segmentation class. Compared to the results in Tab. \ref{tab:segmentation}, all error metrics tend to improve. More specifically, we can notice how the metrics do improve significantly for the most challenging materials, at the cost of a minimal decrease in accuracy within the simpler regions (Class 0). Overall we reckon that, although our experiments show that availability of annotated data can help to better handle specular/transparent objects by deep stereo networks, the accuracy level turns out still much worse compared to opaque surfaces. Hence, we observe that these kinds of materials set forth really hard open challenges  in stereo which, hopefully, may be addressed in future research thanks also to the availability of the annotated data provided by Booster. 

\begin{table}[t]
\centering
\scalebox{0.6}{
\begin{tabular}{ccc}

 \begin{tabular}{c}
    \multirow{3}{*}{\rotatebox{90}{Full res.}} \\ 
 \end{tabular}
 \begin{tabular}{l}
 \\
 \toprule
 \\
 Category \\
 \midrule
 All \\
 \midrule
 Class 0 \\
 Class 1 \\
 Class 2 \\
 Class 3 \\
 \bottomrule
 \end{tabular}
 &
 \begin{tabular}{rrrr | rr }
 \multicolumn{6}{c}{All pixels} \\
 \toprule
 bad-2 & bad-4 & bad-6 & bad-8 & MAE & RMSE \\
 (\%) & (\%) & (\%) & (\%) & (px.) &  (px.) \\
 \midrule
 38.68 & 23.33 & 17.66 & 14.55 & 7.56 & 17.39 \\
 \midrule
 37.50 & 20.47 & 13.75 & 10.40 & 4.43 & 10.07 \\
 42.48 & 23.35 & 16.15 & 12.22 & 5.24 & 12.05 \\
 61.84 & 42.37 & 33.23 & 27.37 & 13.08 & 18.08 \\
 65.59 & 48.74 & 39.19 & 32.93 & 14.91 & 21.75 \\ 
 \bottomrule
 \end{tabular}
 
 \\
 \\

 \begin{tabular}{c}
    \multirow{3}{*}{\rotatebox{90}{Quarter res.}} \\ 
 \end{tabular}
 \begin{tabular}{l}
 \\
 \toprule
 \\
 Category \\
 \midrule
 All \\
 \midrule
 Class 0 \\
 Class 1 \\
 Class 2 \\
 Class 3 \\
 \bottomrule
 \end{tabular}
 &
 \begin{tabular}{rrrr | rr }
 \multicolumn{6}{c}{All pixels} \\
 \toprule
 bad-2 & bad-4 & bad-6 & bad-8 & MAE & RMSE \\
 (\%) & (\%) & (\%) & (\%) & (px.) &  (px.) \\
 \midrule
 14.46 & 9.47 & 7.32 & 5.76 & 1.87 & 4.23 \\
 \midrule
 10.29 & 4.61 & 2.76 & 2.00 & 1.08 & 2.33 \\
 12.09 & 6.35 & 4.62 & 3.58 & 1.28 & 2.82 \\
 27.22 & 16.83 & 13.06 & 10.65 & 3.25 & 4.42 \\
 32.91 & 21.08 & 15.36 & 10.46 & 3.70 & 5.32 \\ 
 \bottomrule
 \end{tabular} \\
 
 \end{tabular}
 }
 \vspace{-0.25cm}\caption{\textbf{Results on the Booster Balanced testing split after fine tuning on the training split  -- material segmentation.} We run RAFT-Stereo, \textbf{fine-tuned} on the Booster training split, processing quarter resolution images. We evaluate on full resolution ground-truth maps, or by downsampling them to quarter resolution.}
 \label{tab:finetuning}
\end{table}

\begin{figure*}[t]
    \centering
    \renewcommand{\tabcolsep}{1pt}
    \begin{tabular}{cccccccc}
        \scriptsize \textit{RGB \& GT} & \scriptsize \textit{MC-CNN} \cite{zbontar2016stereo} & \scriptsize \textit{LEAStereo} \cite{cheng2020hierarchical} & \scriptsize \textit{CFNet} \cite{shen2021cfnet} & \scriptsize \textit{HSMNet} \cite{yang2019hierarchical} & \scriptsize \textit{Neural Ref.} \cite{aleotti2021neural} & \scriptsize \textit{RAFT-Stereo} \cite{lipson2021raft} & \scriptsize \textit{RAFT-Stereo (ft)} \cite{lipson2021raft}\\
        \scriptsize Input Res. & \scriptsize Q & \scriptsize Q & \scriptsize Q & \scriptsize H & \scriptsize H & \scriptsize Q & \scriptsize Q \\    
        
        \includegraphics[width=0.10\textwidth]{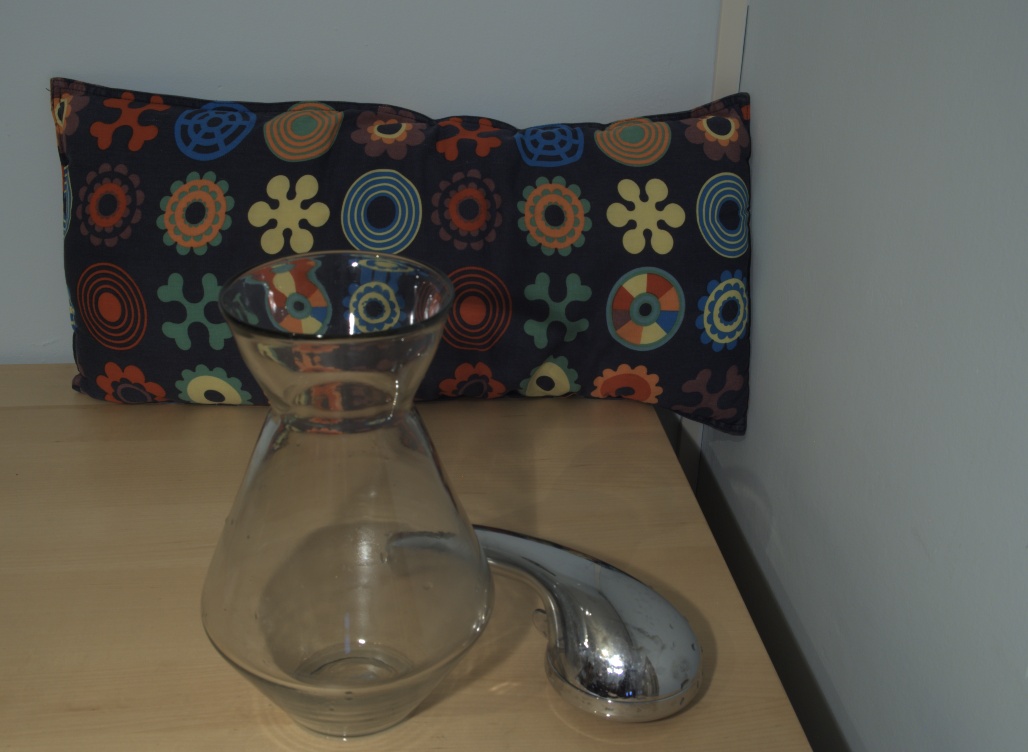} &
        \includegraphics[width=0.10\textwidth]{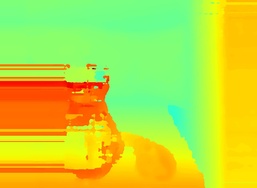} &
        \includegraphics[width=0.10\textwidth]{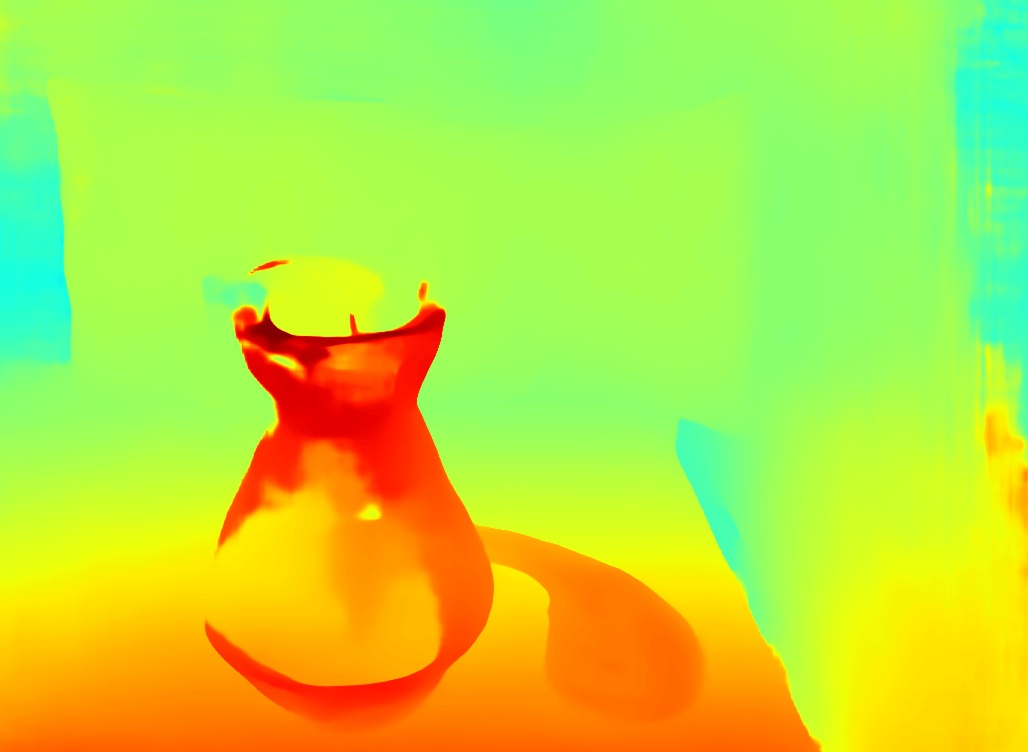} &
        \includegraphics[width=0.10\textwidth]{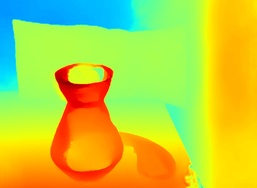} &
        \includegraphics[width=0.10\textwidth]{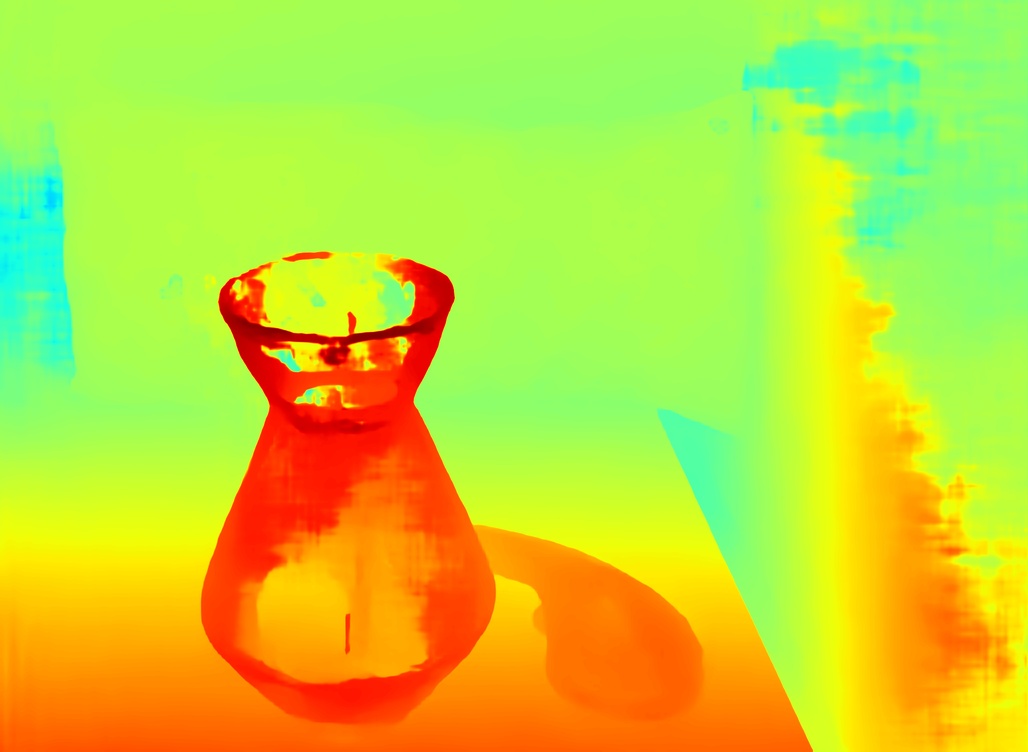} &
        \includegraphics[width=0.10\textwidth]{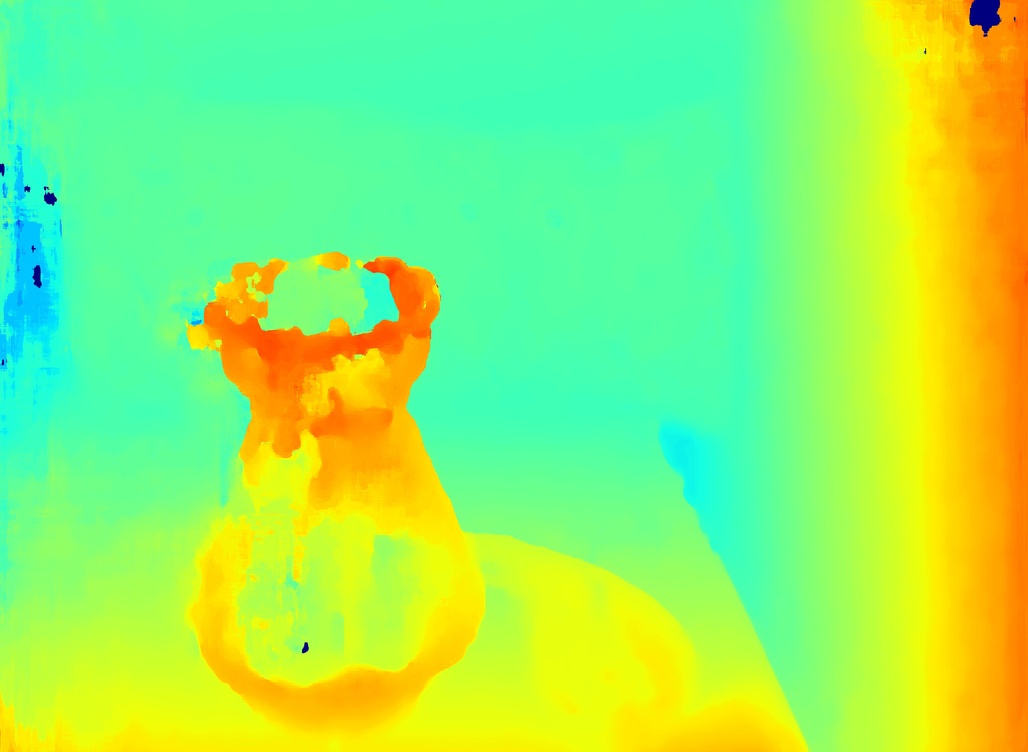} &
        \includegraphics[width=0.10\textwidth]{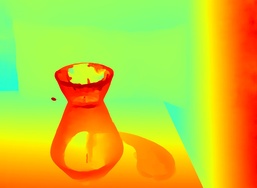} &
        \includegraphics[width=0.10\textwidth]{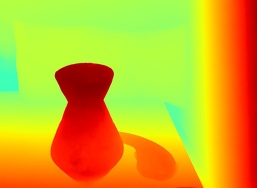} \\
        \includegraphics[width=0.10\textwidth]{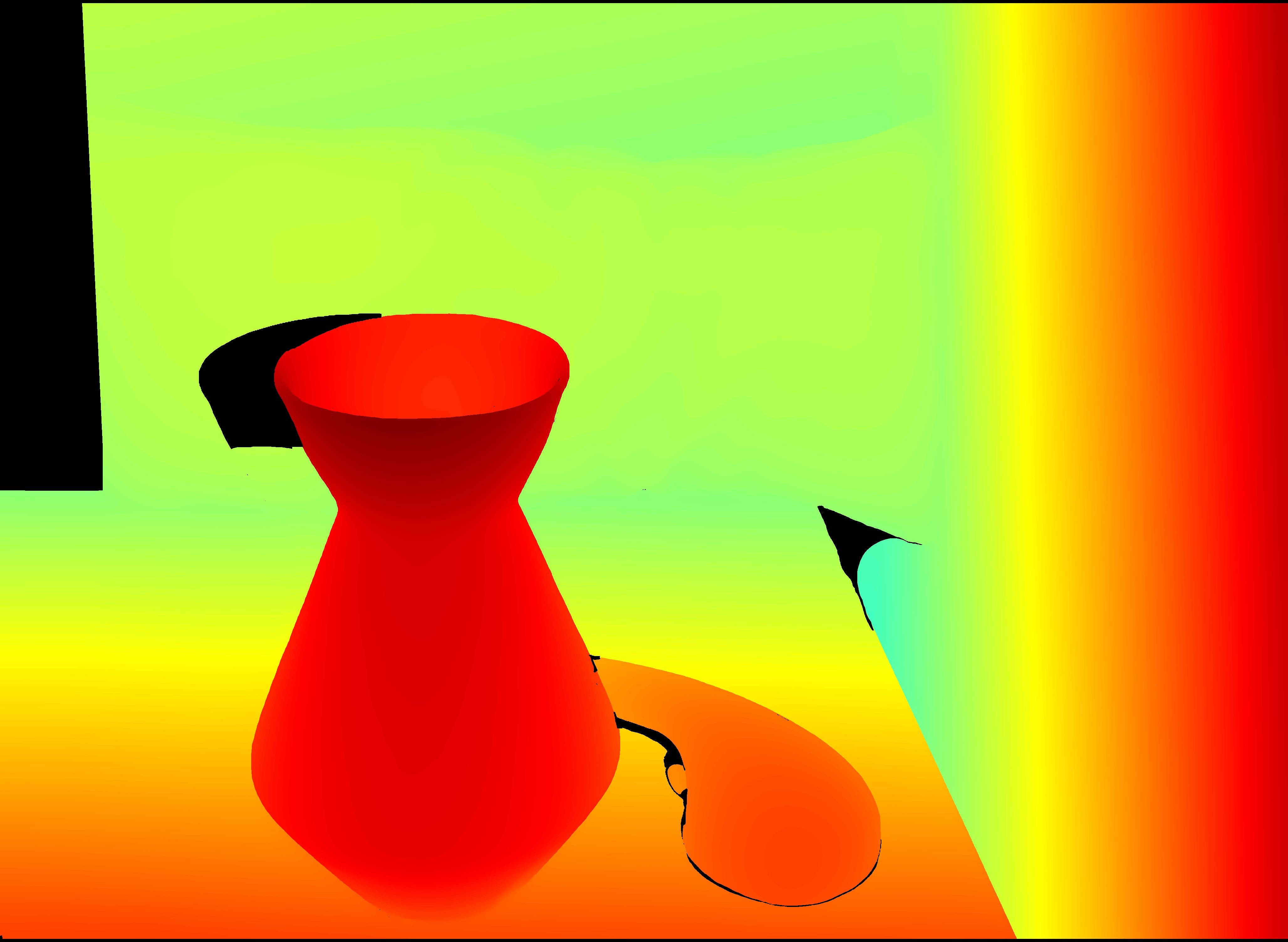} &
        \includegraphics[width=0.10\textwidth]{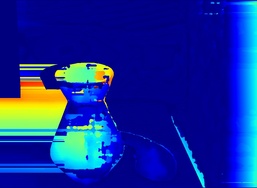} &
        \includegraphics[width=0.10\textwidth]{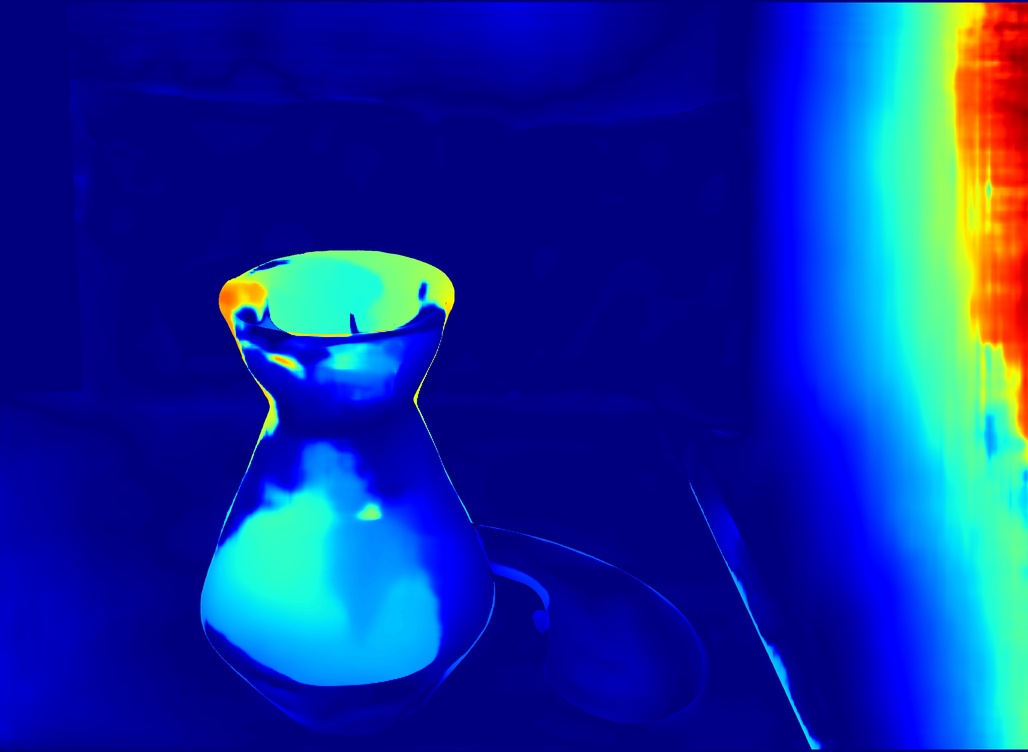} &
        \includegraphics[width=0.10\textwidth]{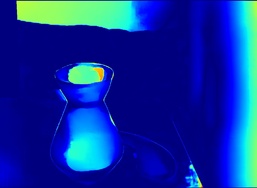} &
        \includegraphics[width=0.10\textwidth]{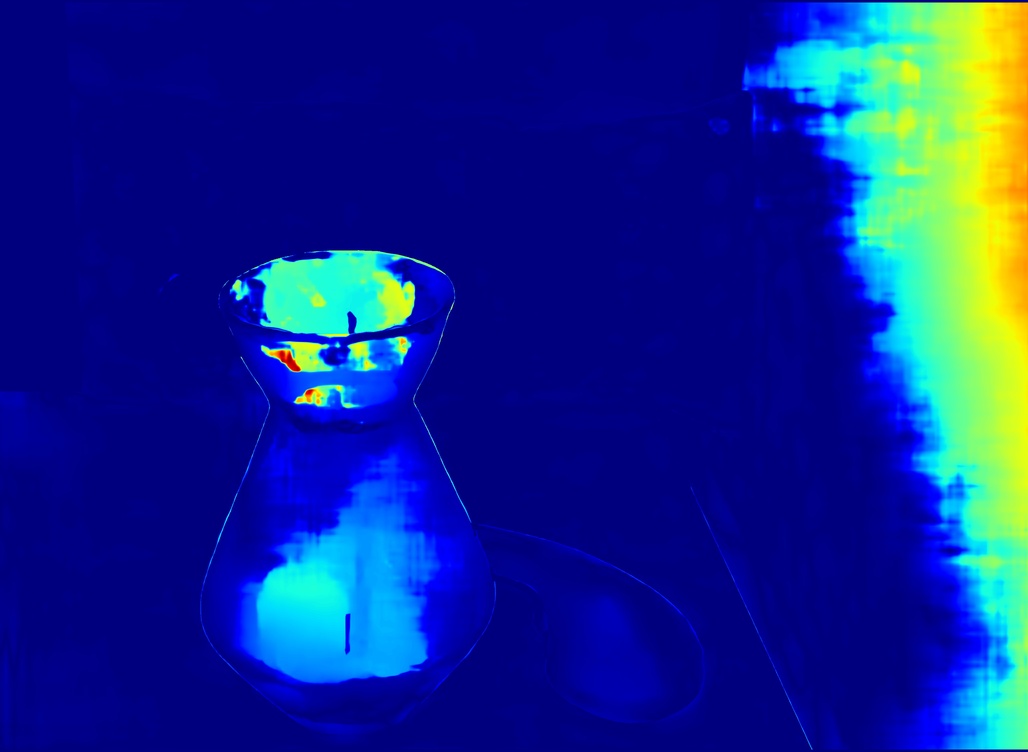} &
        \includegraphics[width=0.10\textwidth]{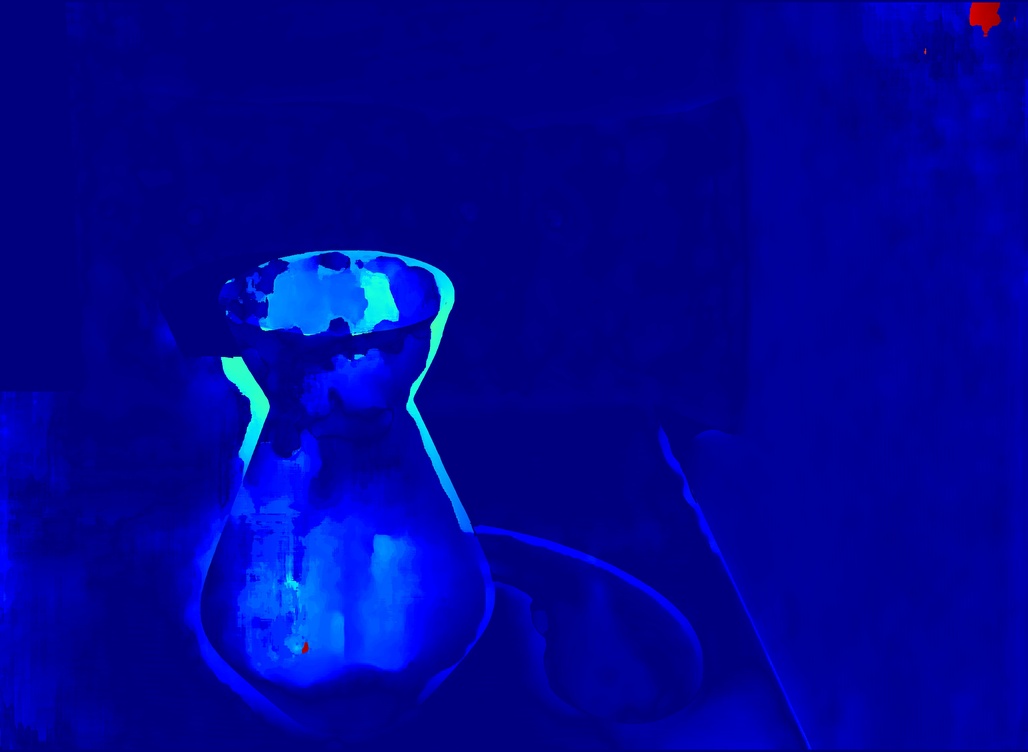} &
        \includegraphics[width=0.10\textwidth]{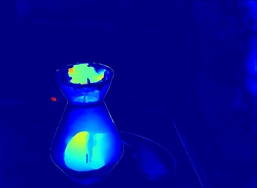} &
        \includegraphics[width=0.10\textwidth]{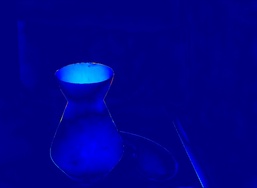} \\
    \end{tabular}
    \vspace{-0.25cm}\caption{\textbf{Qualitative results on Booster Balanced testing split.} We show the reference image (top) and the ground-truth map (bottom) on leftmost column, followed by disparity (top) and error maps (bottom) for the deep models evaluated in our benchmark.}
    \label{fig:tournament}
\end{figure*}

In Fig. \ref{fig:tournament} we provide some qualitative results dealing with  the predictions  obtained by the networks evaluated in Tab. \ref{tab:stereo_tournament} as well as, in the rightmost column, by RAFT-Stereo after fine-tuning on the Booster training set (Tab. \ref{tab:finetuning}). After fine-tuning on the Booster training split, RAFT-Stereo has learned to handle transparent objects much better. 

\subsection{Unbalanced Stereo Benchmark}

Here, we evaluate the considered stereo methods on the Booster Unbalanced testing split.
Tab. \ref{tab:unbalanced} collects the outcome of this experiment. For most methods, we follow the baseline approach defined in \cite{aleotti2021neural} and downsample the reference high-resolution image to the same resolution as the second image. Yet, as HSMNet is designed to handle high-resolution stereo pairs, for this network we upsample the target up to the reference
image size. We point out that these results are not directly comparable to those in Tab. \ref{tab:stereo_tournament}, since the baseline length (and thus disparity values) in this setup is halved, thus making the matching problem easier (i.e., smaller research range). Therefore, being errors larger than those of the Balanced split, it is evident the major difficulty of this scenario.
Moreover, we highlight that, similarly to the Balanced setup, by fine-tuning RAFT-Stereo on the Unbalanced training split, we can improve its performance on nearly all metrics. 
Thus, future research on stereo may leverage the finding that state-of-the-art deep models  hold the potential to better learn to match specular/transparent surfaces even in unbalanced setting when properly fine-tuned with carefully annotated data. 

\begin{table}[t]
\centering
\scalebox{0.65}{
\begin{tabular}{ccc}

 \begin{tabular}{l}
 \\
 \toprule
 \\
 Model \\
 \midrule
 SGM \cite{hirschmuller2007stereo} \\
 MC-CNN \cite{zbontar2016stereo} \\
 LEAStereo \cite{cheng2020hierarchical} \\
 CFNet \cite{shen2021cfnet} \\
 HSMNet \cite{yang2019hierarchical} $\dag$ \\
 SGM+Neural Ref. \cite{aleotti2021neural} $\dag$ \\
 RAFT-Stereo \cite{lipson2021raft} \\
 \midrule
 RAFT-Stereo \cite{lipson2021raft} (ft) \\
 \bottomrule
 \end{tabular}
 &
 \begin{tabular}{rrrr | rr }
 \multicolumn{6}{c}{All pixels} \\
 \toprule
 bad-2 & bad-4 & bad-6 & bad-8 & MAE & RMSE \\
 (\%) & (\%) & (\%) & (\%) & (px.) &  (px.) \\
 \midrule
 78.47 & 62.74 & 52.62 & 45.97 & 42.63 & 97.62 \\ 
 86.30 & 68.67 & 54.20 & 44.78 & 23.64 & 45.46 \\ 
 74.31 & 57.70 & 47.11 & 39.88 & 17.68 & 31.29\\ 
 70.22 & 53.20 & 43.61 & 37.10 & 16.19 & 28.78 \\ 
 63.20 & 43.22 & 32.87 & 26.55 & 11.96 & 22.82 \\ 
 70.90 & 52.15 & 41.71 & 35.40 & 24.27 & 52.52 \\ 
 \bfseries 55.96 & \bfseries 36.81 & \bfseries 27.87 & \bfseries 22.33 & \bfseries 9.86 & \bfseries 19.36 \\ 
 \midrule
 58.67 & 32.83 & 22.96 & 17.65 & 6.31 & 11.11 \\ 
 \bottomrule
 \end{tabular}
 
 \end{tabular}
 }
 \vspace{-0.25cm}\caption{\textbf{Results on Booster Unbalanced testing split.} We run stereo networks, using weights made available by their authors. We evaluate on full resolution ground-truth maps. $\dag$ denotes images being resized to half the reference resolution (about 6 Mpx). (ft) denotes fine-tuned on Booster Unbalanced training split.}
 \label{tab:unbalanced}
\end{table}

\subsection{Challenges in Monocular Depth Estimation} 

\begin{figure}[t]
    \centering
    \renewcommand{\tabcolsep}{1pt}
    \begin{tabular}{cccc}
        \includegraphics[width=0.08\textwidth]{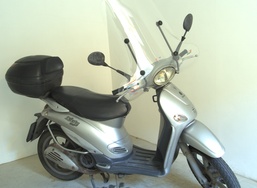} &
        \includegraphics[width=0.08\textwidth]{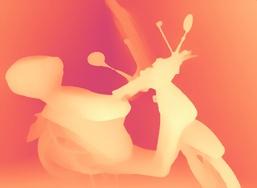} &
        \includegraphics[width=0.08\textwidth]{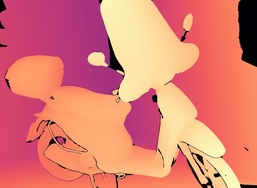} &
        \includegraphics[width=0.08\textwidth]{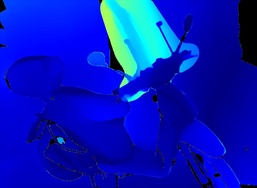} \\
    
        \includegraphics[width=0.08\textwidth]{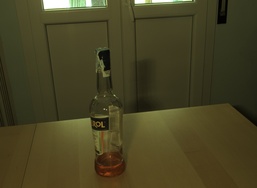} &
        \includegraphics[width=0.08\textwidth]{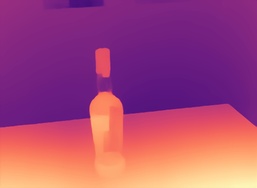} &
        \includegraphics[width=0.08\textwidth]{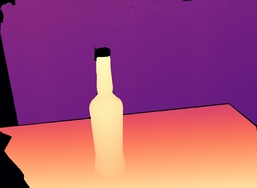} &
        \includegraphics[width=0.08\textwidth]{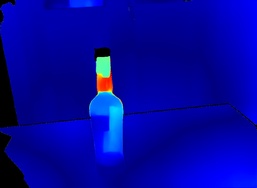} \\
        
        \includegraphics[width=0.08\textwidth]{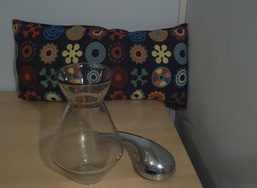} &
        \includegraphics[width=0.08\textwidth]{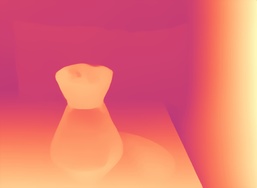} &
        \includegraphics[width=0.08\textwidth]{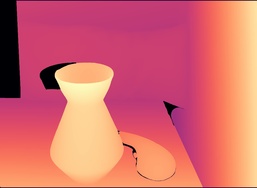} &
        \includegraphics[width=0.08\textwidth]{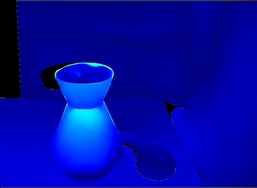} \\
    \end{tabular}
    \vspace{-0.25cm}\caption{\textbf{Qualitative results for monocular depth estimation.} From left to right: reference images, depth maps predictions by DPT \cite{Ranftl_2021_ICCV}, ground-truth depth maps, error maps. }
    \label{fig:mono}
\end{figure}

We argue that most of the difficult surfaces featured by Booster set forth open-challenges to be addressed in future research also for other image-based depth estimation approaches, such as, in particular, monocular depth estimation. As a side experiment, thus, we run DPT \cite{Ranftl_2021_ICCV} -- a transformer for single image depth estimation -- on the images present in our dataset. Fig. \ref{fig:mono} shows some qualitative examples, highlighting how scale-aligned DPT predictions are very inaccurate on transparent surfaces.

\section{Conclusion, Limitations and Future work}
In this paper, we have presented the \underline{B}enchmark \underline{o}n \underline{o}pen-challenges in \underline{ster}eo (\textbf{Booster}), a novel stereo dataset collecting 419 images -- acquired both in \textit{balanced} and \textit{unbalanced} setups -- featuring extremely challenging environments and kinds of objects. It comes with dense and accurate ground-truth disparities, obtained through a novel deep space-time stereo pipeline, as well as with manually annotated material segmentation masks. Compared to recent stereo datasets targeting autonomous/assisted driving, such as DrivingStereo \cite{yang2019drivingstereo}, Booster includes a much smaller number of annotated images and, hence, cannot be considered a \textit{large-scale} dataset. Moreover, the deep space-time pipeline and the small baseline used for annotations constraints the collected scenes to frame indoor environments.

Our experiments show that Booster unveils some of the most intriguing challenges in deep stereo and provides hints on promising research directions. In particular, follow-up work fostered by Booster may be devoted to i) investigating on the ability of deep models suitably fine-tuned on Booster to generalize to outdoor setting featuring similar difficult surfaces and materials, ii) devising a pipeline, e.g, leveraging  on Lidar sensors, to collect annotated data with transparent/specular surfaces also in outdoor setting, iii) build large scale synthetic datasets specifically addressing the open-challenges highlighted by Booster to enable more effective pre-training and iv) building
and scanning scenes through successive depth layers, to gather multiple depths at transparent/reflective objects which can be useful for applications such as augmented reality.

Therefore, we are lead to believe that Booster holds the potential to  \textit{boost} future research in deep stereo.

\textbf{Acknowledgements.} We gratefully acknowledge the funding support of Huawei Technologies Oy (Finland). We thank Andrea Pumilia for helping collecting/cleaning data.

{\small
\bibliographystyle{ieee_fullname}
\bibliography{egbib}
}

\newpage\phantom{Supplementary}
\multido{\i=1+1}{8}{


\section*{Supplementary material}

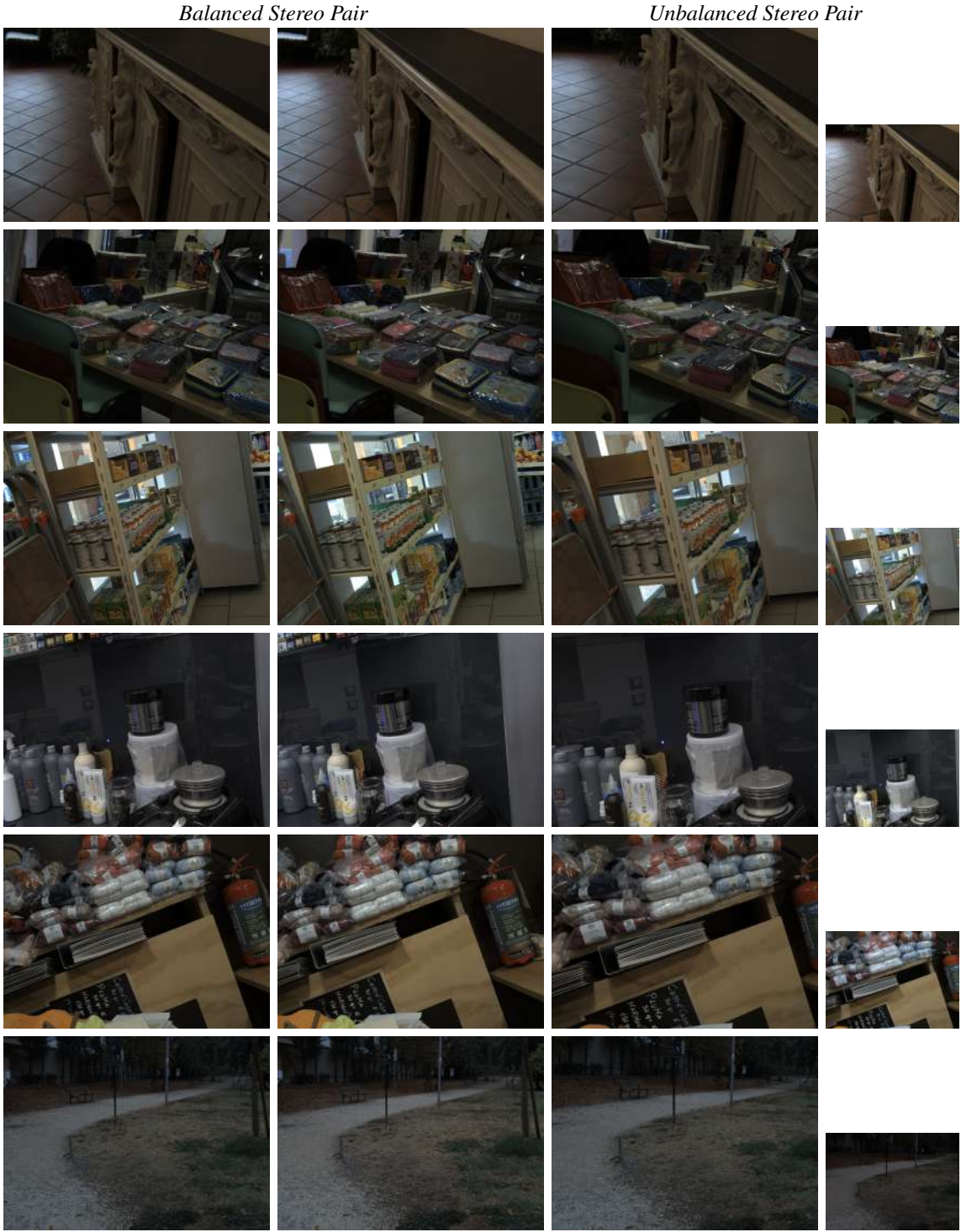

Figure 6. **Booster passive examples.** From left to right: left balanced, right balanced, left unbalanced, right unbalanced images.

## 6. Additional Details on Disparity Warping

As outlined in Sec. 3 of the main paper, in the case of the unbalanced $L-C$ stereo system, we need to warp the left ground-truth $Disp_L^{LR}$ aligned with $L-R$ to obtain the ground-truth $Disp_L^{LC}$ aligned with the left image of $L-C$.

We know that the rectification transformation is only a

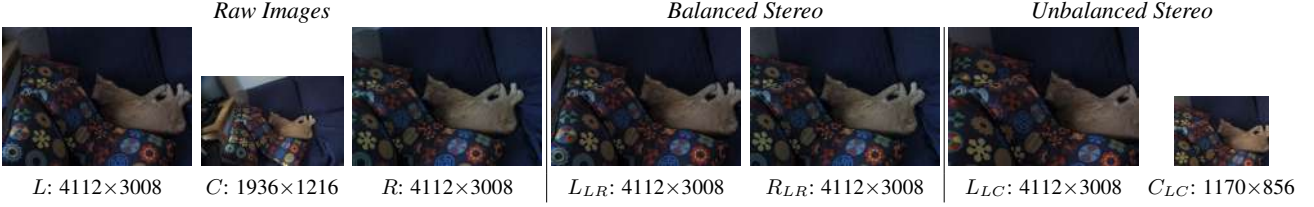

Figure 7. **Rectification example.** From left to right: $L$, $R$, and $C$ raw images acquired by our trinocular rig. $L_{LR}$, $R_{LR}$ rectified balanced stereo pair from of the $L-R$ stereo system. $L_{LC}$, $C_{LC}$ unbalanced rectified stereo pair from of the $L-C$ stereo system.

change of intrinsic parameters and a rotation, thus an homography. Therefore, we can calculate the mapping between pixels of the left image $L_{LR}$ of the $L-R$ stereo system, with coordinate $(u,v)$, and pixels of the left image $L_{LC}$ of the $L-C$ stereo system, with coordinate $(u',v')$ as:

$$\begin{pmatrix} u' \\ v' \\ 1' \end{pmatrix} = A_L^{LC} R_L^{LC} R_L^{LR^{-1}} A_L^{LR^{-1}} \begin{pmatrix} u \\ v \\ 1 \end{pmatrix} \quad (5)$$

Known this mapping we can perform a backward warping to obtain $Disp_L^{LC}$ from $Disp_L^{LR}$. However, we need to be change the disparity values according to the 3D rotation and baseline change before warping. Thus, given the disparity map $Disp_L^{LR}$ we first transform it to the corresponding depth map $D_L^{LR}$ as follows:

$$D_L^{LR} = \frac{f_{LR} b_{LR}}{Disp_L^{LR}} \quad (6)$$

Where $f_{LR}$ is the focal length of the $L_{LR}$ and $b_{LR}$ is the baseline of the stereo system $L-R$. Then, we back-project each pixel of $L_{LR}$ to 3D using $D_L^{LR}$ and we rotate it accordingly to $R$ obtaining the pixel in the $L_{LC}$ reference frame:

$$\begin{pmatrix} x' \\ y' \\ z' \end{pmatrix} = R_L^{LC} R_L^{LR^{-1}} D_L^{LR} A_L^{LR^{-1}} \begin{pmatrix} u \\ v \\ 1 \end{pmatrix} \quad (7)$$

In this way we can create a depth map $D_L^{LR \to LC}$ for which any pixel $(u,v)$ contains the depth value of the corresponding pixel aligned in the $L_{LC}$ reference frame, $z'$. At this point we perform the backward warping of the depth:

$$D_L^{LC} = \phi(D_L^{LR \to LC}) \quad (8)$$

where $\phi$ is the backward warping operation that use the mapping defined at Eq. 5 and $D_L^{LC}$ is the depth map aligned with $L_{LC}$. Finally we transform it to the ground disparity map of $L_{LC}$ as:

$$Disp_L^{LC} = \frac{f_{LC} b_{LC}}{D_L^{LC}} \quad (9)$$

where $f_{LC}$ and $b_{LC}$ are the focal length of $L_{LC}$ and the baseline of the $L-C$ stereo system.


}

\end{document}